\documentclass[screen, nonacm]{acmart}
\AtBeginDocument{%
  }

\usepackage{subcaption} 
\usepackage{tikz}
\usetikzlibrary{arrows.meta, positioning, calc}
\usepackage{amsmath}
\usepackage{graphicx}
\usepackage{booktabs}
\usepackage{tabularx}
\usepackage[table,xcdraw]{xcolor}
\usepackage{siunitx}
\usepackage{multirow}
\usepackage{siunitx}
\usepackage{colortbl}
\definecolor{dm_blue}{HTML}{E8F0FE}
\usepackage{array}         
\usepackage{pifont}        
\usepackage{threeparttable} 

\usepackage{soul}
\usepackage{xcolor}

\newcommand{\cmark}{\ding{51}}
\newcommand{\xmark}{\ding{55}}

\begin{document}

\title{Tide: A Customisable Dataset Generator for Anti-Money Laundering Research}

\author{Montijn van den Beukel}
\email{montijnvandenbeukel@gmail.com}
\affiliation{%
  \institution{University of Amsterdam}
  \city{Amsterdam}
  \country{The Netherlands}}

\author{Jo\v{z}e M. Ro\v{z}anec}\thanks{Corresponding author: Jo\v{z}e M. Ro\v{z}anec: joze.rozanec@ijs.si}
\email{joze.rozanec@ijs.si}
\affiliation{%
  \institution{Jo\v{z}ef Stefan Institute}
  \city{Ljubljana}
  \country{Slovenia}}

\author{Ana-Lucia Varbanescu}
\email{a.l.varbanescu@utwente.nl}
\affiliation{%
  \institution{University of Amsterdam/University of Twente}
  \city{Amsterdam}
  \country{The Netherlands}}

\renewcommand{\shortauthors}{van den Beukel et al.}

\begin{abstract}
The lack of accessible transactional data significantly hinders machine learning research for Anti-Money Laundering (AML). Privacy and legal concerns prevent the sharing of real financial data, while existing synthetic generators focus on simplistic structural patterns and neglect the temporal dynamics (timing and frequency) that characterise sophisticated laundering schemes.

We present Tide, an open-source synthetic dataset generator that produces graph-based financial networks incorporating money laundering patterns defined by both structural and temporal characteristics. Tide enables reproducible, customisable dataset generation tailored to specific research needs. We release two reference datasets with varying illicit ratios (LI: 0.10\%, HI: 0.19\%), alongside the implementation of state-of-the-art detection models.

Evaluation across these datasets reveals condition-dependent model rankings: LightGBM achieves the highest PR-AUC (78.05) in the low illicit ratio condition, while XGBoost performs best (85.12) at higher fraud prevalence. These divergent rankings demonstrate that the reference datasets can meaningfully differentiate model capabilities across operational conditions.

Tide provides the research community with a configurable benchmark that exposes meaningful performance variation across model architectures, advancing the development of robust AML detection methods.
\end{abstract}





\settopmatter{printacmref=false, printccs=false, printfolios=true}
\renewcommand\footnotetextcopyrightpermission[1]{}
\maketitle

\section{Introduction}

Money laundering enables criminals to disguise illegal funds by routing them through complex transactions to integrate them into the legitimate financial system. These funds often fuel other illicit activities, including drug distribution, bribery, corruption, and terrorism~\cite{levi2006money}. An estimated 2–5\% of global GDP is laundered annually, amounting to \$800 billion to \$2 trillion, though the clandestine nature of these operations makes precise measurement difficult~\cite{unodc_money_laundering_overview}. The fight against money laundering imposes a significant financial burden: in the United States alone, financial crime compliance costs exceeded \$59 billion in 2023~\cite{forrester2023truecost}, with anti-money laundering (AML) efforts accounting for \$4.8 billion to \$8 billion annually, yet yielding fewer than 700 convictions~\cite{burton2021proposals}. The challenge of detecting illicit schemes buried within vast amounts of legitimate financial traffic remains substantial, as most money laundering goes undetected and regulatory definitions of suspicious activity remain poorly defined~\cite{ustreasury2022nmlra, bello2017uncertainty}.


\subsection{Machine Learning for Fraud Detection}
Financial transactions naturally form complex networks in which entities are connected by monetary flows; thus, they can be represented as directed graphs. This graph-based representation has proven valuable for money laundering detection, as illicit activities often manifest as distinctive network patterns, such as fan-in, fan-out, scatter-gather, and cyclic structures, that are difficult to detect when analysing transactions in isolation~\cite{savage2016moneylaundering_groups, altman2023realistic}.

While traditional machine learning approaches such as Support Vector Machines, Random Forests, and Neural Networks have been applied to financial fraud detection~\cite{soria2024ml_aml_review}, these methods typically operate on tabular transaction features without explicitly modelling relational structure among entities. Graph Neural Networks (GNNs) address this limitation by learning directly from network topology alongside transaction attributes, and have shown promising results for anti-money laundering and related financial crime detection tasks~\cite{motie2023financial, deprez2025continual}. Yet regardless of the approach chosen, developing robust detection models remains difficult due to challenges that extend beyond model architecture.

\subsection{Current Limitations and Challenges}
These challenges are primarily rooted in data access and availability. First, legal, privacy, and competition concerns prevent financial institutions from sharing real transaction data~\cite{altman2023realistic}, limiting access to high-quality training datasets. Second, studies presenting classification models suffer from poor reproducibility due to a lack of accessible datasets; researchers cannot replicate experiments, validate claims, or benchmark new methods against existing work. While synthetic datasets offer a promising solution by providing shareable alternatives that represent real-world characteristics, existing generators have significant limitations. They often ignore the temporal dimensions of laundering patterns, focusing solely on structural relationships~\cite{shadrooh2024smotef}. Furthermore, they lack fine-grained customisability for adjusting entity attributes and pattern parameters, and implement only basic structural patterns while omitting more sophisticated real-world schemes~\cite{austrac_typologies_2019}.

\subsection{Contributions}
To address these limitations, we present Tide, a customisable synthetic data generator that produces reproducible datasets modelling both the structural and temporal aspects of money laundering patterns. At the broadest level, Tide addresses the shortage of accessible data that has limited progress in money laundering detection research. Our contributions are fourfold: (1) we introduce the Tide generator, which advances synthetic data generation by incorporating temporal dynamics alongside structural relationships, offering extensive customisability for graph parameters and pattern characteristics; (2) we release reference datasets to facilitate immediate reproducibility and allow researchers to benchmark detection methods against consistent data; (3) we provide implementations of state-of-the-art detection models, serving both to validate the synthetic data and to establish robust baselines for future AML research; and (4) we perform a comprehensive performance analysis of the generator, demonstrating its scalability and efficiency in generating large-scale networks.

\subsection{Paper Organization}
The remainder of this paper is structured as follows. Section~\ref{sec:related} reviews related work on fraud characterization, existing dataset generators, and money laundering detection models. Section~\ref{sec:methodology} presents our research methodology, including pattern identification, and our graph modelling approach. Section~\ref{sec:generation} describes Tide's implementation in detail, including our reference dataset. Section~\ref{sec:evaluation} presents our evaluation results, demonstrating both structural validity and machine learning model performance on the generated datasets. Finally, Section~\ref{sec:conclusion} summarises our contributions, discusses limitations, and outlines future work.

\section{Related work}
\label{sec:related}
While the financial burden of AML compliance is immense, the operational failure of legacy systems stems primarily from their reliance on deterministic logic. Historically, institutions relied on Transaction Monitoring Systems (TMS) defined by static rules and thresholds. These systems are inherently reactive, flagging activities only when they deviate from prescribed norms, such as transfers exceeding a fixed value or interactions with high-risk jurisdictions. Consequently, they generate false positive rates often exceeding 95\%, overwhelming analysts and diluting the focus required to identify genuine financial crime~\cite{richardson_williams_mikkelsen2019_networkaml}.

Machine learning methods aim to provide more precise identification by moving beyond static thresholds to learn adaptive decision boundaries. Among them, unsupervised methods address legacy limitations by modelling normal behaviour and identifying anomalies. Clustering algorithms like DBSCAN group entities with similar transaction profiles, flagging outliers for investigation~\cite{yang2014dbscan_aml, ester1996density}. However, validation remains challenging without ground truth labels, and traditional approaches often analyse transactions in isolation rather than capturing coordinated network activities.

\subsection{Money Laundering Detection Models}

\subsubsection{Traditional Machine Learning Methods}
To overcome the rigidity of rule-based systems, machine learning approaches have been widely adopted for financial fraud detection. A systematic review by Soria et al.~\cite{soria2024ml_aml_review} examined models including Support Vector Machines (SVMs), Random Forests, k-Nearest Neighbours, and Artificial Neural Networks, finding that SVMs achieved the highest reported accuracy among traditional approaches. Gradient Boosted Trees, particularly XGBoost, have become an industry standard for tabular financial data due to their ability to learn non-linear patterns, handle extreme class imbalance, and provide interpretable baselines for benchmarking~\cite{jensen2022ml_vs_rules_aml}.

However, these methods suffer from a fundamental limitation: they treat transactions as independent, isolated events. They cannot inherently capture the relational dependencies—such as circular flows or long chains of intermediaries—that characterise sophisticated money laundering schemes, often requiring extensive feature engineering to approximate network structure.

\subsubsection{Graph Representation Learning}
Graph-based supervised learning has emerged as a promising approach for financial fraud detection, with growing adoption in recent years~\cite{motie2023financial, deprez2025continual}. By representing financial systems as directed graphs, with entities as nodes and transactions as edges, these methods can model complex, high-dimensional relational patterns that traditional approaches cannot capture~\cite{he2021efficient}. Graph Neural Networks (GNNs), including Graph Convolutional Networks (GCNs)~\cite{alarab2020gcn_aml_bitcoin}, Graph Attention Networks (GATs)~\cite{velickovic2018graph_attention}, and Graph Isomorphism Networks (GINs)~\cite{xu2019how_powerful_gnn}, learn from both node features and network structure through neighbourhood aggregation, enabling the detection of sophisticated patterns such as circular fund flows~\cite{deprez2025continual}.

However, standard GNNs operate under the assumption of \textit{homophily}, the principle that connected nodes share similar labels. Financial fraud networks, conversely, are typically \textit{heterophilic}: illicit actors strategically connect with legitimate accounts to camouflage their activities~\cite{zhu2021gnn_heterophily}. In such settings, standard aggregation mechanisms smooth out the distinct features of fraud nodes by mixing them with the dominant benign signals from their neighbours, often degrading detection performance~\cite{zhu2021gnn_heterophily}. While recent architectures like CPGNN attempt to model this heterophily by learning a compatibility matrix~\cite{zhu2021gnn_heterophily}, they require dense, representative training data to learn these complex class correlations.

Furthermore, the representational power of these models is strictly limited by the topological complexity of available training data. Without datasets that exhibit both the intricate structural-temporal dependencies and the heterophilic nature of real-world laundering, GNNs cannot be trained or benchmarked robustly.

\subsubsection{Class Imbalance in Fraud Detection}
A defining characteristic of financial fraud datasets is extreme class imbalance, in which legitimate transactions substantially outnumber fraudulent ones~\cite{makki2019imbalanced_cc_fraud}. This imbalance poses significant challenges for classification algorithms: due to the dominance of the benign class, optimisation steps tend to correctly classify the majority while ignoring the minority, causing fraudulent transactions to go undetected~\cite{makki2019imbalanced_cc_fraud}. In graph-based settings, this problem is further compounded during message passing, where the influence exerted by the vast majority of benign neighbours overwhelms the signal from fraudulent connections, and gradient updates become dominated by the majority class~\cite{zhuo2024partitioning}. Consequently, the model's ability to learn discriminative patterns for identifying illicit edges is compromised. While prior work has attempted to mitigate this through techniques such as oversampling (e.g., SMOTE) or cost-sensitive learning, these approaches often yield increased false positive rates without substantially improving fraud detection~\cite{makki2019imbalanced_cc_fraud}, underscoring the need for benchmarks that simulate realistic fraud patterns at scale.

\subsection{Fraud Characterization and Pattern Requirements}
\label{backgr:patterns}

Money laundering typically follows a three-stage process: placement (introducing illicit funds into the financial system),
layering (obscuring the origin through complex transactions), and integration (re-entering funds into the legitimate
economy) \cite{singh2019amlviz}. However, detecting sophisticated schemes requires analyzing the interdependence between \textit{structural characteristics} (network topology) and \textit{temporal characteristics} (timing and frequency) \cite{tariq2023topology}. Existing literature has identified specific typologies that exhibit these features. We categorize these patterns below to establish the functional requirements for our generator.

\subsubsection{Temporal Dynamics}
\label{ref:temporal_patterns}
Certain patterns are defined primarily by transaction timing rather than unique topological structures. High-Frequency Transactions involve sudden bursts of activity or rapid flow-through that deviate from an entity's established history, often involving cash structuring below reporting thresholds \cite{li2020flowscope, austrac2014typologies}. Similarly, synchronised transactions imply collusion, where seemingly unrelated entities transact (e.g., cash deposits) at nearly the same time to coordinate a larger transfer \cite{shadrooh2024smotef}. Periodic Transactions attempt to mimic legitimate salary or bill payments but lack a clear economic purpose for the involved entities \cite{austrac2014typologies}.

\subsubsection{Network Topology Structures}
\label{ref:network_topology}
Graph-based detection often seeks specific subgraph patterns. Star Topologies feature a central node (mule or shell company) interacting with multiple outer nodes, characterized by high in-degree or out-degree centrality \cite{he2021efficient}. Cliques represent fully connected subgraphs; $k$-cliques (where $k \ge 4$) with bimodal edge curvature distributions are strong indicators of coordinated rings rather than random chance \cite{granados2022geometry}. Front Businesses introduce dense connectivity among business accounts, whereby cash deposits are immediately converted into international transfers \cite{irwin2012modelling}.

\subsubsection{Transaction Chains}
Sophisticated layering involves extended transaction chains. U-Turn Transactions create international loops in which funds leave and return to the originator's jurisdiction, obscuring the trail \cite{singh2019amlviz}. Low-Degree Path Hopping and Parallel Structures (or Smurfing) involve splitting funds across long chains of intermediaries or parallel paths to dilute the value of individual transactions before consolidation \cite{ he2021efficient}.

\subsection{Synthetic Data Generation Methodologies}
The legal and privacy barriers surrounding real financial data have necessitated a reliance on synthetic generation within the Anti-Money Laundering (AML) research community \cite{altman2023realistic, jensen2023synthetic}. To contextualize our contribution, we draw on recent systematic reviews that survey the landscape of synthetic financial data. In an analysis of 72 studies published since 2018, Meldrum et al. \cite{meldrum2025syntheticdata_finance} identify a heavy concentration of research on stock market time-series, with transaction data appearing in only a small fraction of works. This scarcity is corroborated by Motie and Raahemi \cite{motie2023financial}, whose systematic review of Graph Neural Networks (GNNs) for fraud detection highlights the limited availability of diverse graph datasets across financial domains and identifies the study of dynamic graphs as a critical research gap. Similarly, a 2025 review by Wilson and Azmani \cite{wilson2025_gan_finance} confirms that, although Generative Adversarial Networks (GANs) have become widely adopted for financial data generation, significant challenges persist, including mode collapse, training instability, and difficulty capturing competitive market dynamics such as strategic interactions between market participants.

Current approaches can be broadly categorized into two paradigms: \textit{Process-Driven} (Agent-Based) methods, which simulate economic behavior from the bottom up, and \textit{Data-Driven} (Generative) methods, which learn statistical distributions from seed data.

\subsubsection{Process-Driven Approaches (Agent-Based Modeling)}
Process-driven methods circumvent the need for sensitive training data by relying on predefined logic and typologies.
AMLSim \cite{AMLSim} remains among the most widely used generators for this approach. While it successfully generates valid graph structures with injected patterns, it focuses primarily on the layering stage and lacks the mechanisms to inject the precise temporal signatures required to benchmark modern Temporal Graph Networks (TGNs).

AMLWorld \cite{altman2023realistic} improves upon this by employing a ``Virtual World'' model, where agent actions are governed by strict, calibrated schedules (e.g., weekly salaries, monthly interest payments) rather than simple stochastic interactions, thereby capturing temporal regularity in both legitimate and illicit flows. However, unlike AMLSim, AMLWorld is distributed primarily as a static benchmark dataset rather than an accessible generation tool. This creates a barrier for researchers who require custom scenarios, as they are restricted to the fixed set of laundering patterns (e.g., fan-in, scatter-gather) pre-baked into the released data, with no mechanism to programmatically inject novel or evolving typologies.

The importance of such structural flexibility is underscored by Singh and Best \cite{singh2019amlviz}, who demonstrate that money laundering manifests as distinct topological ``fingerprints'': specific subgraph patterns that investigators must isolate from background noise. This observation highlights a critical limitation in current simulation tools: to effectively train and benchmark detection systems, a generator must be able to explicitly and deterministically inject known topological patterns, rather than relying on stochastic agents to produce them by chance.

\subsubsection{Data-Driven Approaches (Generative AI \& Statistical)}
Data-driven methods attempt to clone the statistical properties of real datasets. As highlighted by recent surveys, GANs are widely adopted for generative techniques in finance \cite{wilson2025_gan_finance}. Architectures such as CTGAN \cite{xu2019modeling} have shown success in reproducing tabular and time-series distributions. However, as Wilson and Azmani \cite{wilson2025_gan_finance} note, these models struggle with topological integrity in the graph domain and remain prone to mode collapse when generating diverse transaction patterns. They typically treat transactions as independent tabular rows, failing to preserve the complex degree distributions of financial networks.

A notable exception to GAN based approaches is SynthAML \cite{jensen2023synthetic}, which employs a statistical approach that uses Gaussian copulas trained on real banking data. While SynthAML achieves high distributional fidelity to its source, it inherently suffers from low customizability: it is constrained to the specific view of the single institution used for training (seed data) and cannot simulate the complex, cross-border flows essential to international AML research.

\begin{table*}[t!]
    \centering
    \caption{Comparison of synthetic financial transaction generators. Unlike predecessors that offer either fixed patterns (AMLSim) or statistical approximations (SynthAML), \textbf{Tide} provides a dual capability: it is preloaded with a library of complex, well-researched typologies (e.g., U-Turns, Smurfing) while simultaneously allowing the programmatic injection of novel, user-defined patterns.}
    \label{tab:generator_comparison}
    \small
    \renewcommand{\arraystretch}{1.3}
    \begin{tabular}{l c c c c}
        \toprule
        \textbf{Feature} & \textbf{AMLSim} \cite{AMLSim} & \textbf{AMLWorld} \cite{altman2023realistic} & \textbf{SynthAML} \cite{jensen2023synthetic} & \textbf{Tide (Ours)} \\
        \midrule
        \textbf{Core Approach} & Agent-Based & Virtual World Agents & Statistical (Copulas) & Pattern Injection \\
        \midrule
        Executable Generator & \cmark & \xmark & \xmark & \cmark \\
        Pre-loaded Pattern Library & \cmark \textsuperscript{1} & \cmark \textsuperscript{1} & \xmark & \cmark \textsuperscript{2} \\
        Custom Pattern Injection & \xmark & \xmark & \xmark & \cmark \\
        Precise Temporal Control & \xmark & \xmark & \xmark & \cmark \\
        Zero-Shot (No Real Data) & \cmark & \cmark & \xmark & \cmark \\
        \bottomrule
    \end{tabular}
    \vspace{0.5em}
    
    \begin{tablenotes}
        \small
        \item \textsuperscript{1} \textbf{Basic Library}: Limited to standard, structural-only shapes.
        \item \textsuperscript{2} \textbf{Complex Library}: Includes advanced spatio-temporal schemes derived from literature.
    \end{tablenotes}
\end{table*}

\subsubsection{The Gap: Interdependence of Structure and Time}
A synthesis of these reviews reveals a clear dichotomy. Process-driven tools are structurally valid but operationally limited: either by the rigidity of the codebase (AMLSim) or the unavailability of the generator itself (AMLWorld). Conversely, data-driven tools (GANs) offer statistical realism but suffer from mode collapse and a lack of control over specific typological injections \cite{wilson2025_gan_finance}.

Motie and Raahemi \cite{motie2023financial} conclude that tackling the dynamic, time-evolving nature of fraud graphs is a critical future direction for the field. \textit{Tide} addresses this gap by providing an open, executable framework that enables the explicit injection of patterns defined by both network topologies and precise temporal constraints, thereby bridging the divide between structural validity and temporal complexity. Consequently, it achieves this by providing a library of complex typologies found in the literature \cite{austrac_typologies_2019, austrac2014typologies, irwin2012modelling, he2021efficient, jfcu_fiu_typologies_2023} while supporting the fully custom injection of user-defined constraints.

\section{Research Methodology}
\label{sec:methodology}

\subsection{Overview}
\begin{figure}[ht]
    \centering
    \begin{tikzpicture}[
        node distance=1.5cm and 1.5cm,
        auto,
        block/.style={
            rectangle, 
            draw=black, 
            thick, 
            fill=blue!5, 
            text width=3.0cm, 
            align=center, 
            rounded corners, 
            minimum height=1.6cm,
            font=\small
        },
        cloud/.style={
            draw=red!50, 
            thick, 
            dashed, 
            fill=red!5, 
            text width=2.8cm, 
            align=center, 
            minimum height=1.2cm,
            rounded corners,
            font=\small\bfseries
        },
        line/.style={
            draw, 
            thick, 
            -{Latex[length=2.5mm, width=2.5mm]}, 
            shorten >=2pt
        }
    ]

    
    \node [block] (research) {\textbf{1. Pattern}\\ \textbf{Research}\\ \footnotesize(Typology Abstraction)};

    \node [block, right=1.2cm of research] (design) {\textbf{2. Generator}\\ \textbf{Design}\\ \footnotesize(Graph Modelling)};

    \node [block, below=1.2cm of design] (gen) {\textbf{3. Synthetic}\\ \textbf{Generation}\\ \footnotesize(Data Production)};

    \node [block, below=1.2cm of research] (eval) {\textbf{4. Adversarial}\\ \textbf{Evaluation}\\ \footnotesize(ML Detection Tests)};

    \node [cloud, left=1.2cm of eval] (final) {Final\\ Benchmarking\\ \footnotesize(Paper Results)};


    \path [line] (research) -- (design);
    
    \path [line] (design) -- (gen);

    \path [line] (gen) -- (eval);

    \path [line] (eval.north east) -- node [midway, above, sloped, font=\footnotesize] {\textit{Calibrate parameters}} (design.south west);

    \path [line] (eval) -- (final);

    \end{tikzpicture}
    \caption{The iterative research methodology. The process employs an adversarial feedback loop where generator parameters (Phase 2) are recalibrated based on the performance of detection models (Phase 4).}
    \label{fig:methodology_cycle}
\end{figure}
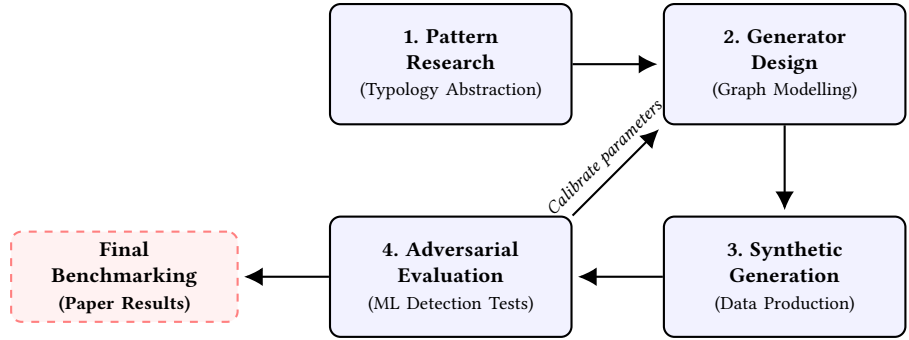

When performing this research, we adopted an iterative, feedback-driven development lifecycle. As illustrated in Figure~\ref{fig:methodology_cycle}, the process consisted of four phases structured around an optimization loop:

\begin{enumerate}
    \item \textit{Typological Abstraction}: We created injection rules by abstracting structural and temporal constraints from regulatory literature and AML case studies.
    \item \textit{Generative Framework Design}: These rules were formalized into the Tide generator, which injected specific fraud topologies into a background of statistically calibrated legitimate activity.
    \item \textit{Optimization Loop}: This phase constituted the core refinement cycle. We generated provisional datasets and subject them to a diverse committee of baseline detection models. We utilize the Youden Index \cite{fluss2005estimation} ($J$) to determine the decision threshold for each classifier, ensuring that performance is benchmarked at the point that maximizes the separation between legitimate and illicit distributions.
    \item \textit{Calibration \& Refinement}: Based on the detection feedback, we adjusted the generator's difficulty parameters while strictly enforcing statistical invariants. The Area Under the Precision-Recall Curve (PR-AUC) served as our primary difficulty gauge, as it is robust to the extreme class imbalance inherent in financial fraud. If patterns were detected with trivial ease (e.g., PR-AUC $> 0.90$), we could introduce higher temporal variance or topological camouflage; if they were undetectable (approaching the random baseline), we could relax the constraints. This cycle was repeated until the datasets exhibited realistic detection complexity that challenges classifiers while remaining solvable.
\end{enumerate}

Once the generator is calibrated, we proceed to the final comparative benchmarking (Section~\ref{sec:results}), where the tool is evaluated against state-of-the-art baselines.

\subsection{Data Modelling and Typology Formalization}
To ensure the generated datasets accurately reflect the complexity of financial crime, we first established a robust data model. We represented the financial domain as a heterogeneous, directed graph comprising multiple entities. Following this modelling phase, we conducted a review of regulatory reports \cite{austrac2014typologies, austrac_typologies_2019} and academic literature \cite{jfcu_fiu_typologies_2023, granados2022geometry, li2020flowscope, shadrooh2024smotef, singh2019amlviz} to identify recurring laundering typologies.

We then mapped these qualitative descriptions into executable constraints. To guarantee the topological integrity of the identified schemes, we generate the fraud schemes in isolation. In this hierarchical process, distinct laundering patterns are instantiated first to strictly enforce their structural and temporal requirements. Subsequently, a background generation process simulates general financial traffic, enveloping the fraudulent entities in noise. To formalize these descriptions, we represent each money-laundering pattern as a tuple $P = (S, T)$:
\begin{itemize}
    \item \textbf{The Structural Component ($S$)} defines the criteria for selecting the entities that will participate in the pattern instance. It acts as a query to find nodes with specific attributes from the pre-computed clusters.
    \item \textbf{The Temporal Component ($T$)} defines the sequence of transactions between the selected entities, including their timing, amounts, and other edge attributes.
\end{itemize}

\subsection{Validation Framework and Iteration}
We define a two-stage validation process to assess the quality and utility of the generated datasets.

\paragraph{Structural Validation}
This phase verifies that the generated graph topology and transaction patterns strictly adhere to their formal definitions and user configurations. It confirms that the selected entities satisfy pattern-specific criteria and that transaction sequences comply with the defined temporal constraints.

\paragraph{Machine Learning Model Performance Evaluation}
This phase assesses the dataset's practical utility by training a selection of detection models on the generated data. The objective is twofold: to ensure that embedded patterns are realistic enough to be detectable yet subtle enough to pose a non-trivial challenge for money laundering detection. Additionally, this process established a feedback loop in which early experimental results guided refinements to the generation logic to enhance data realism.

\subsection{Performance Assessment Metrics}
\label{sec:performance_metrics}
To evaluate machine learning model performance on the generated datasets, we employ standard binary classification metrics that facilitate direct comparison with established benchmarks, specifically AMLWorld~\cite{altman2023realistic}, while including supplementary measures for robustness. To ensure reported metrics reflect each model's maximum potential effectiveness rather than an arbitrary probability cut-off, we determine classification thresholds using Youden's $J$ statistic (Equation~\ref{eq:youden}), which identifies the point that maximizes separation between the legitimate and illicit class distributions. This threshold-selection approach makes performance metrics invariant to class prevalence, providing a fair comparison across datasets with varying imbalance ratios.
\begin{equation}
\label{eq:youden}
J = \text{Sensitivity} + \text{Specificity} - 1 = \text{Recall} + (1 - \text{FPR}) - 1
\end{equation}

The following metrics are reported:
\begin{itemize}
    \item \textit{F1-Score (at Youden Index):} The harmonic mean of precision and recall, adopted as the primary metric to align with AMLWorld benchmarks.
    \item \textit{Precision (at Youden Index):} The proportion of flagged transactions that are truly illicit, reflecting investigative efficiency. High precision minimizes false positives, reducing the operational burden of reviewing legitimate transactions.
    \item \textit{Recall (at Youden Index):} The proportion of illicit transactions successfully identified, capturing detection coverage. High recall ensures fewer illicit flows escape detection.
    \item \textit{PR-AUC:} Provides rigorous assessment under extreme class imbalance, sensitive to false positives, and reflecting the operational cost of false alarms.
\end{itemize}

\section{Synthetic Dataset Generation}
\label{sec:generation}


This section presents Tide, our synthetic data generator designed to address the scarcity of high-quality, accessible datasets for Anti-Money Laundering (AML) research. Unlike existing generators that often prioritize network topology over precise transaction sequences, Tide explicitly models both the structural and temporal characteristics of financial crime. The system creates directed, heterogeneous graphs containing both legitimate economic activity and injected fraud patterns, providing a ground truth for training and benchmarking supervised detection models.

\subsection{Guiding Principles and Design Goals}
\label{sec:design_goals}
To effectively model these laundering typologies, the architecture of Tide was developed to meet four functional requirements. These goals address specific limitations found in existing synthetic data tools:

\paragraph{Temporal Representation} The generator must model the temporal aspects of money laundering. Many schemes are defined by the sequence, timing, and frequency of transactions. Therefore, the system needs to support time-ordered interactions.

\paragraph{Complex Structural Patterns} The generator needs to create patterns based on both structural and temporal characteristics. The implementation has to account for the attributes of the entities involved, allowing patterns to be defined by a combination of who is transacting and how their profiles contribute to the scheme's logic.

\paragraph{Customisability} To be a useful research tool, the generator needs to be highly configurable. This includes control over the graph's overall size and fraud ratio, the specific parameters of each laundering pattern, and the characteristics of the legitimate background activity.

\paragraph{Reproducibility} To facilitate benchmark comparisons, the generation process must be deterministic, producing identical datasets when provided with the same configuration and random seed.

\subsection{Graph Data Model}
\label{sec:graph_model}
We model the financial network as a directed, heterogeneous graph $G = (V, E)$. This structure supports distinct entity types and directed financial flows.

\subsubsection{Entity and Attribute Design}
The nodes \( v \in V \) in the graph represent four distinct entity types: \textit{individuals}, \textit{businesses}, \textit{accounts}, and \textit{financial institutions}. Each node has a set of common attributes (Table~\ref{tab:common_node_attributes}) and type-specific attributes that define its characteristics (Tables \ref{tab:individual_node_attributes}-\ref{tab:fi_node_attributes} in the Appendix).

\begin{table}[h!]
\centering
\begin{tabularx}{\textwidth}{l X l}
\toprule
\textbf{Attribute} & \textbf{Description} & \textbf{Type} \\
\midrule
\textbf{node\_id} & Unique identifier for each node. & String \\
\textbf{node\_type} & The type of the entity. & Categorical \\
\textbf{country\_code} & Country where the entity is based. & String \\
\textbf{is\_fraudulent} & Label indicating whether node is part of a fraud pattern. & Boolean \\
\textbf{risk\_score} & Value in [0, 0.9] expressing the probability of an entity being fraudulent. & Float \\
\bottomrule
\end{tabularx}
\caption{Common attributes for all node types.}
\label{tab:common_node_attributes}
\end{table}
To model cash-based activities, two mechanisms were implemented. First, each individual and business has a dedicated `cash` account for withdrawals. Second, a special global `cash` node ($v_{\text{cash}}$) acts as a source for cash deposits where the origin is unknown.

\paragraph{Risk Scoring Framework}
To systematically model the likelihood of an entity's involvement in illicit activity, a risk scoring function was developed. This risk-scoring function is used to select entities for fraudulent patterns in the later stage of the generation process. Note that risk scores are excluded from the feature set provided to machine learning models during training and evaluation, ensuring that detection performance reflects learned patterns rather than direct access to ground-truth selection criteria. For individuals and businesses, the score is calculated as a weighted combination of known risk factors:
\begin{equation}
\label{eq:risk_score}
\text{risk}(v) = \min\left(\text{base risk} + \sum_{i} w_i \cdot f_i(v), 0.9\right)
\end{equation}
In this function, \( w_i \) is the weight for risk factor \( i \), and \( f_i(v) \) is an indicator function (1 if the factor is present, 0 otherwise). The default weights were calibrated based on their reported prevalence in literature~\cite{austrac2014typologies}, with factors like operating a cash-intensive business or being located in a high-risk jurisdiction assigned higher weights (See Table~\ref{tab:risk-weights}).

\paragraph{Entity Clustering}
\label{sub:clustering}
To support efficient entity selection for pattern injection in a large graph, we define a clustering function that groups individual and business entities based on their risk profiles:

\begin{equation}
\text{cluster}: V_{\text{entities}} \rightarrow 2^{\mathcal{C}} \setminus \{\emptyset\}
\end{equation}

where $V_{\text{entities}} \subseteq V$ contains only individuals and businesses (excluding accounts and institutions), and $\mathcal{C}$ represents a predefined set of clusters which consists of both single clusters (single risk factors such as geographic, demographic, occupational, and behavioural factors) and composite clusters that combine specific risk factors. Entities are by default assigned to the `legit' cluster (and removed if they are part of a fraud pattern), so each entity is assigned to at least one cluster, and can belong to multiple clusters simultaneously. This clustering lets Tide select entities that match specific pattern requirements.

\subsubsection{Edge Representation}
Edges \( e \in E \) represent the interactions between nodes. They are defined as a tuple \( e = (i, j, r, a, t, \delta, \gamma, \lambda) \), where each component captures a key piece of information about the interaction (see Table~\ref{tab:edge_tuple}). This detailed structure allows for modelling both static ownership relationships and time-sensitive transaction events.

\begin{table}[htb!]
\centering
\begin{tabularx}{\textwidth}{l X l}
\toprule
\textbf{Component} & \textbf{Description} & \textbf{Type} \\
\midrule
\textbf{$i, j$} & Source and target node IDs. & String \\
\textbf{$r$} & Relationship type: \texttt{transaction} or \texttt{ownership}. & Categorical \\
\textbf{$a$} & Transaction amount. & Float \\
\textbf{$t$} & Timestamp of the interaction. & Timestamp \\
\textbf{$\delta$} & Time since the source node's last transaction. & Timedelta \\
\textbf{$\gamma$} & Transaction category. & Categorical \\
\textbf{$\lambda$} & Fraud flag. & Boolean \\
\bottomrule
\end{tabularx}
\caption{Attributes of an edge tuple $e = (i, j, r, a, t, \delta, \gamma, \lambda)$.}
\label{tab:edge_tuple}
\end{table}

\subsection{Generator Architecture and Workflow}
\label{sec:methodology_operational_flow}
\begin{figure}[htbp!]
    \centering
    \subfloat[Step 1: Different entities are created, and then clustered based on attributes.]{\includegraphics[width=0.48\textwidth]{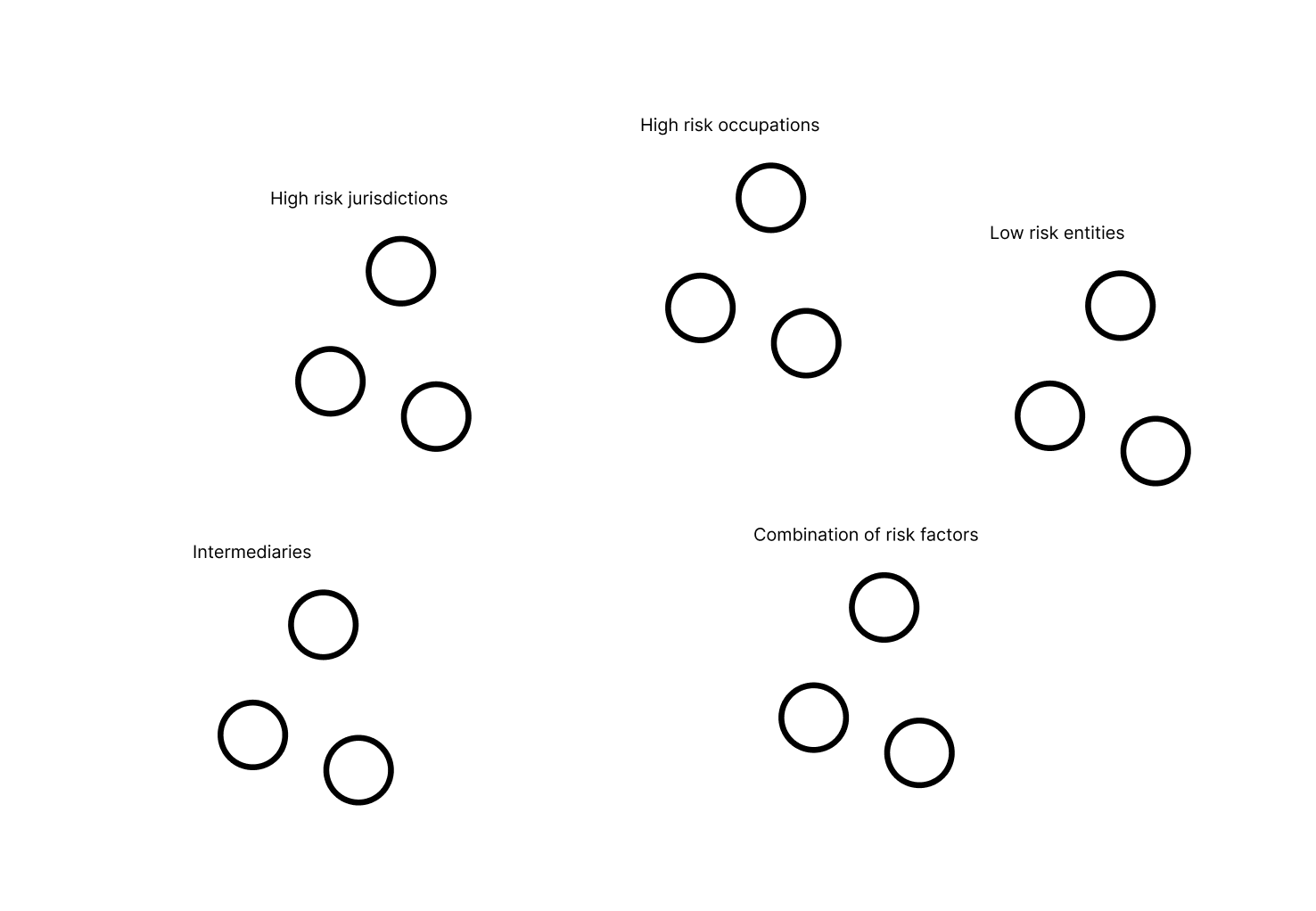}\label{fig:tide_step1_detail}}
    \hfill
    \subfloat[Step 2: Entities are selected for a pattern; the source actor is highlighted in red.]{\includegraphics[width=0.48\textwidth]{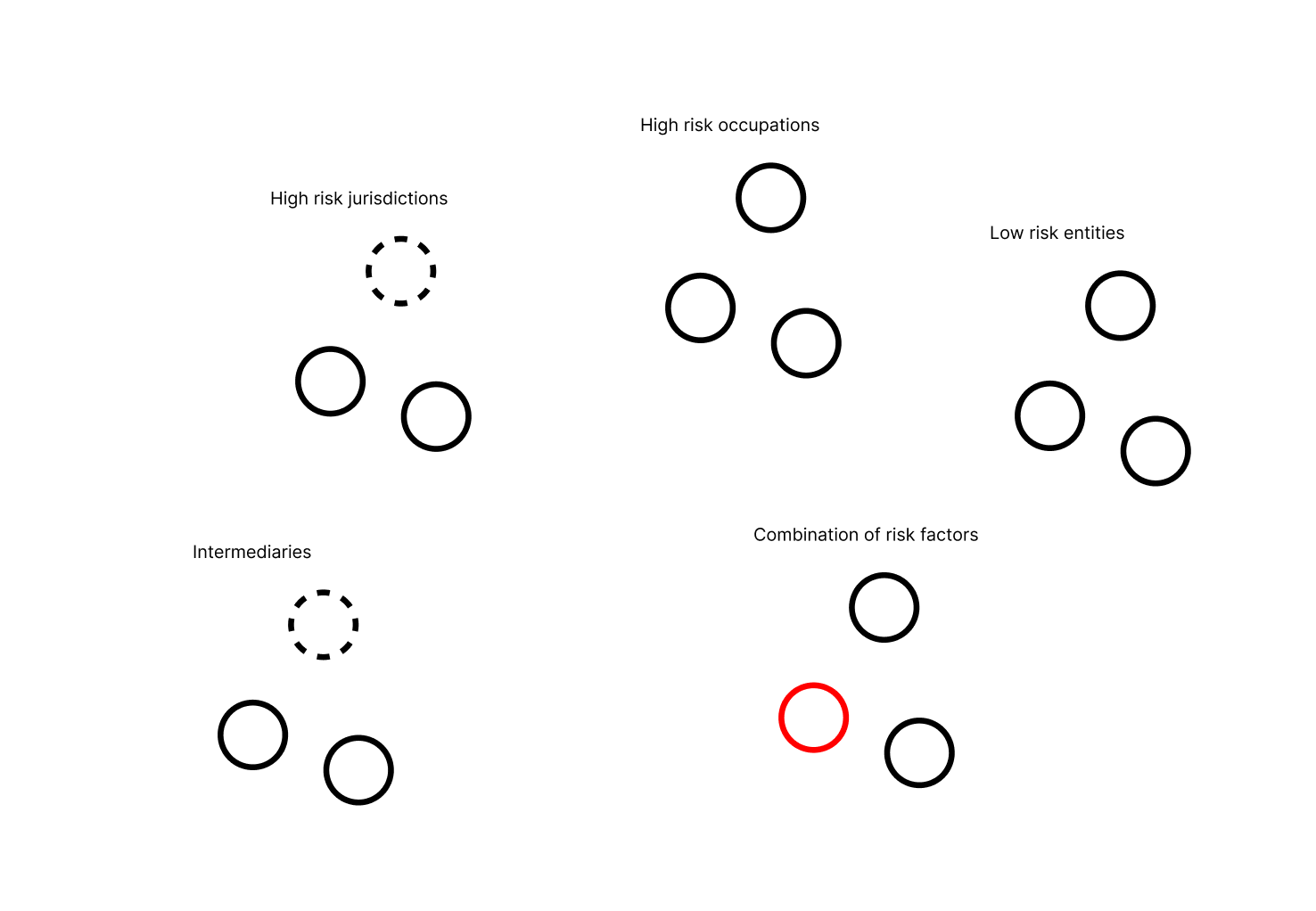}\label{fig:tide_step2_detail}}
    \\ 
    \subfloat[Step 3: Transactions sequences are created between selected entities.]{\includegraphics[width=0.48\textwidth]{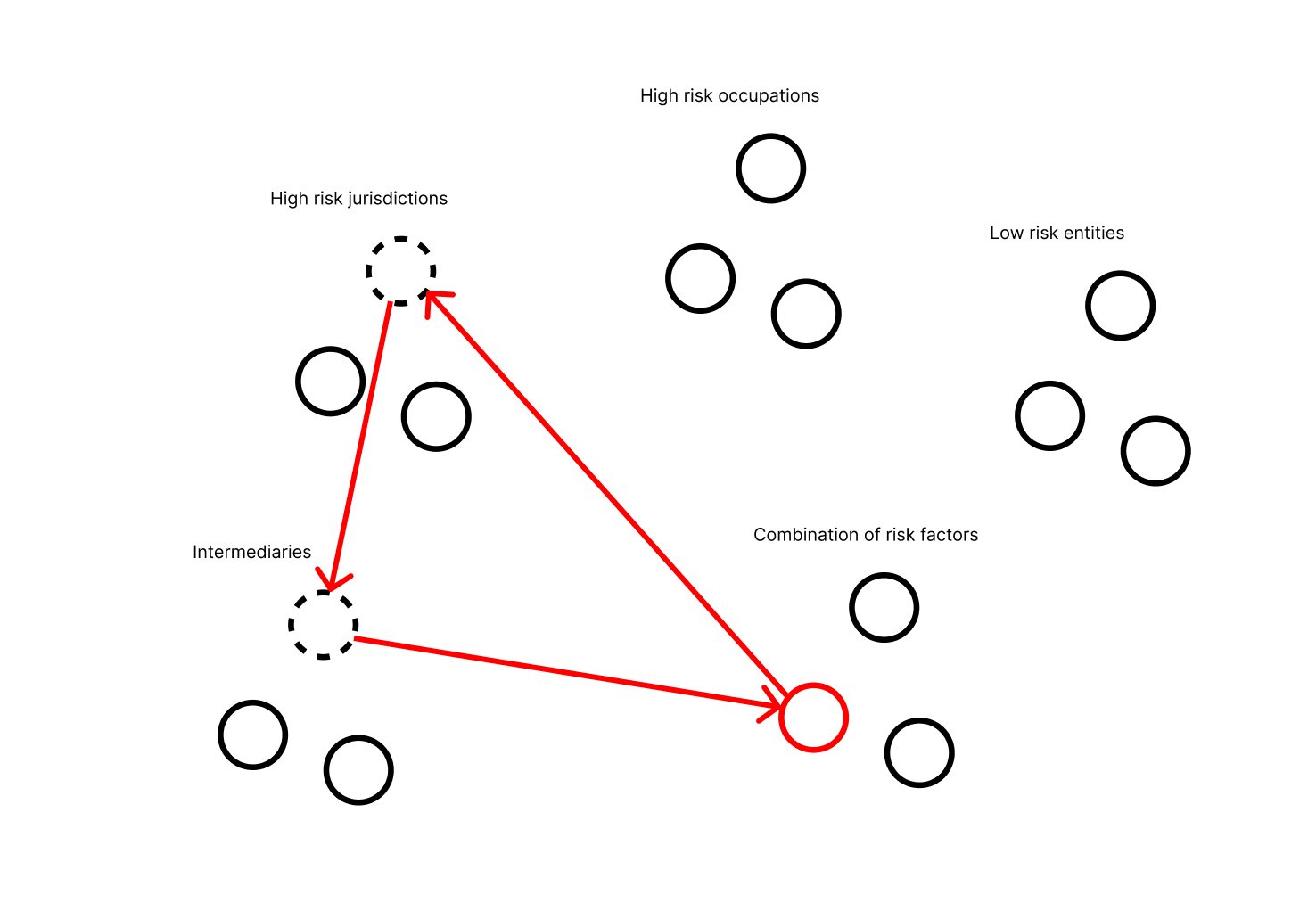}\label{fig:tide_step3_detail}}
    \hfill
    \subfloat[Step 4: Multiple patterns and background activity are merged into a single graph.]{\includegraphics[width=0.48\textwidth]{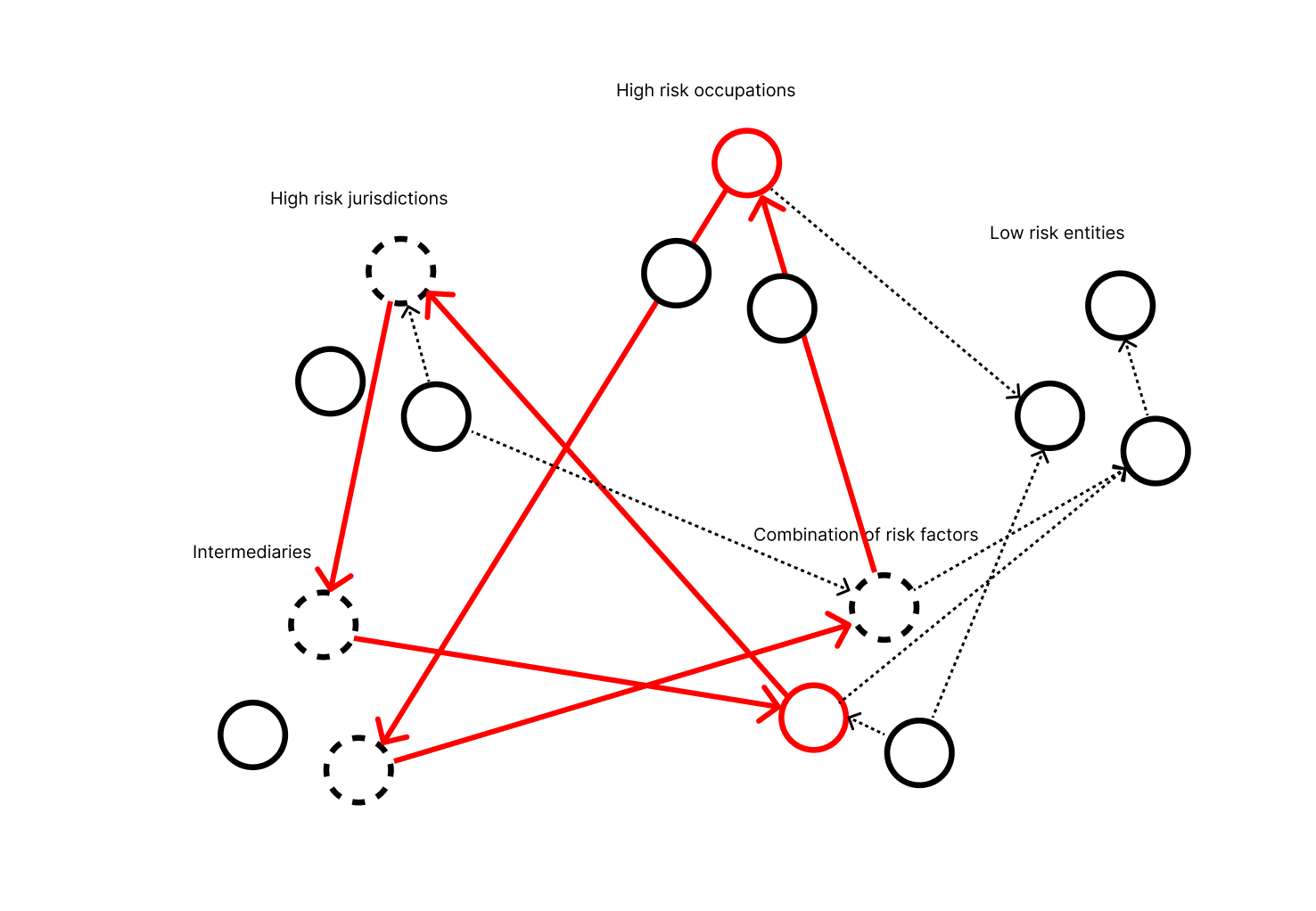}\label{fig:tide_step4_detail}}
    \caption{The four main steps of the Tide graph generation process: (a) Entity creation and clustering, (b) Entity selection, (c) Transaction sequence generation, and (d) Pattern aggregation.}
    \label{fig:tide_generation_steps}
\end{figure}

Tide uses a generation algorithm to construct the synthetic financial graph. The core logic of this pattern injection is illustrated in Figure~\ref{fig:tide_generation_steps} and described below. Note that the fraud pattern depicted in the figure is a simplified representation intended to visualise the concept.

\begin{enumerate}
    \item \textbf{Entity Creation and Clustering (Figure~\ref{fig:tide_step1_detail}):}
    The process begins by populating the graph with financial entities, which are individuals, businesses, accounts, and financial institutions. Each entity is created with a set of attributes as defined in Section~\ref{sec:graph_model}. After this, these entities are grouped into various clusters based on their characteristics (Section~\ref{sub:clustering}). These clusters are used for quick targeted selection of nodes in the following pattern injection phases, so they can operate in near-constant time relative to the total graph size. Low-risk entities form the general population of the graph.

    \item \textbf{Entity Selection for Patterns (Figure~\ref{fig:tide_step2_detail}):}
    For each specific money laundering pattern that needs to be injected into the graph, the generator first identifies the roles required by the pattern's structure. For example, the repeated overseas transfer needs a high risk source (because of their occupation, business, country, or age), with multiple destination nodes in high risk jurisdictions overseas.

    \item \textbf{Transaction Sequence Generation (Figure~\ref{fig:tide_step3_detail}):}
    Once the entities for a pattern instance are selected, the temporal component of that pattern generates a sequence of financial transactions between them. This involves defining the timing, frequency, amounts, and types of these transactions to reflect the specific laundering pattern being modelled. For instance, funds might flow from the primary actor to an intermediary, and then from the intermediary to an entity in a high-risk jurisdiction, as depicted by the red arrows.

    \item \textbf{Pattern Aggregation (Figure~\ref{fig:tide_step4_detail}):}
    The process of selecting entities and generating transaction sequences (steps 2 and 3) is repeated for every money laundering pattern instance that the user configures to be injected. Each pattern generates its own set of fraudulent transactions. These individual pattern sub-graphs are then merged into the main graph. Alongside these illicit activities, legitimate background transactions (represented by black dotted arrows) are also simulated. The final output is a single graph containing both legitimate and fraudulent activities.
\end{enumerate}

\subsection{Fraudulent Pattern Injection}
\label{sec:fraud_patterns}
This section details the money laundering typologies that Tide injects during steps 2 and 3 of the generation workflow (Section~\ref{sec:methodology_operational_flow}). Each pattern defines a structural component, specifying which entities to select from the pre-computed clusters and how they relate, as well as a temporal component governing transaction timing and amounts. Table~\ref{tab:pattern_overview} summarises the five implemented typologies.

\paragraph{Temporal profiles}
We distinguish two temporal profiles for transaction sequences $\{e_1, ..., e_n\}$ with timestamps $\{t_1, ..., t_n\}$: (1) \textit{burst} patterns, where all transactions occur within a bounded window ($\max(t_i - t_j) \leq \Delta_{\text{burst}}$), and (2) \textit{periodic} patterns, where transactions recur at regular intervals ($|t_{i+1} - t_i - \Delta_{\text{period}}| \leq \epsilon$).

\subsubsection{Layering}
\label{sec:layering}
Four of the five patterns optionally insert intermediate layering hops between the source and destination of each fraudulent transfer.
When enabled, each transfer is routed through $h$ randomly selected intermediary accounts, where $h \in [h_{\min}, h_{\max}]$ is drawn uniformly (by default $h \in [2,5]$).
At each hop, the amount decays by a factor $\delta \sim \mathcal{U}(0.95, 0.99)$ to simulate fees or conversion losses, and a delay of 1--48 hours is added between consecutive hops.
Intermediary accounts are selected either uniformly at random from the existing account population, or from a clustered pool of high-risk entities to increase edge-label homophily.
The only pattern that does not use this mechanism is the Synchronised Transactions pattern, whose structure involves direct deposits to a common recipient.
All layering parameters are configurable; disabling layering reduces each pattern to its base two-party topology.

\subsubsection{Repeated overseas transfers}
This pattern models the transfer pattern, where individuals frequently send large transfers to overseas entities.

\begin{figure}[h!]
    \centering
    \includegraphics[width=0.5\linewidth]{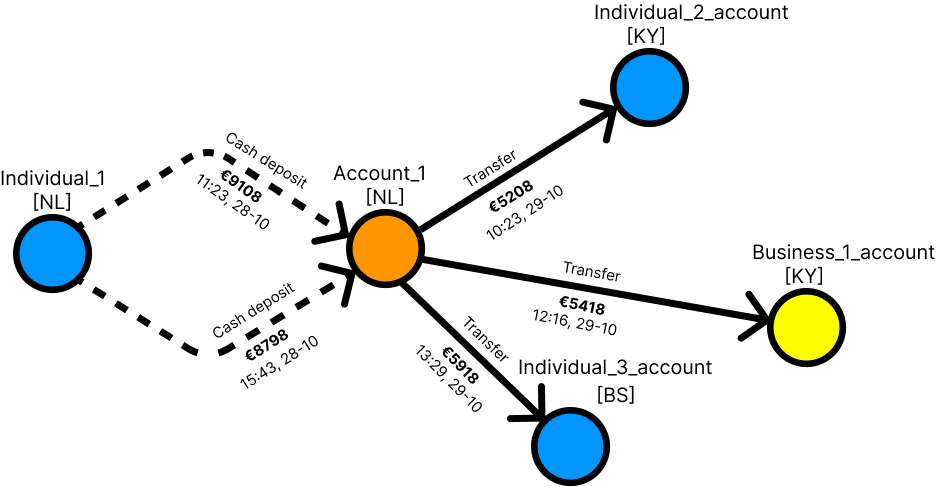}
    \caption{Visualisation of the overseas transfer pattern. \texttt{Individual\_1} owns \texttt{Account\_1}.}
    \label{fig:overseas}
\end{figure}

\paragraph{Structural component}
Tide identifies a source individual from a high-risk cluster. The entity should have at least one domestic bank account. It then selects multiple international destination accounts. High-risk jurisdictions have the highest priority in the destination selection, but other overseas accounts can be selected if no entities are available in the cluster.

\paragraph{Temporal component}
The implementation of this pattern consists of two different behavioural profiles observed in the literature. The first is a burst pattern, where transactions occur within a small timeframe or with small intervals between transactions, defined by a parameterised distribution to avoid mechanical-like timing. The second is a periodic pattern, where transactions occur at regular intervals such as weekly or monthly over a longer time period, as described in Section~\ref{ref:temporal_patterns}. The periodic variant mimics legitimate recurring payments but differs in destination, amount, or lack of clear economic purpose compared to the entity's profile.

The source entity first gets cash deposits under the user-defined reporting threshold, which are then, shortly thereafter, sent to the overseas entities.

\subsubsection{Rapid fund movement}
\label{meth:rapidmovement}
In this pattern, accounts receive high volumes of incoming transfers followed by frequent cash withdrawals.

\begin{figure}[h!]
    \centering
    \includegraphics[width=0.5\linewidth]{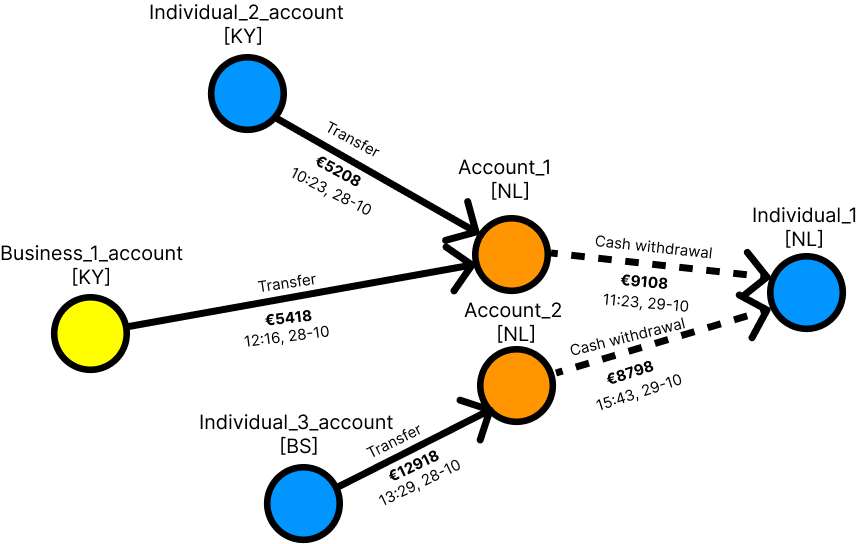}
    \caption{Visualisation of the rapid fund movement pattern. \texttt{Individual\_1} owns \texttt{Account\_1} and \texttt{Account\_2}.}
    \label{fig:rapidflow}
\end{figure}

\paragraph{Structural component}
Tide selects a high-risk individual that owns multiple bank accounts. This is the individual that receives the funds and withdraws money. It also selects multiple overseas entities, again prioritised from high-risk jurisdictions.

\paragraph{Temporal component}
The implementation of this pattern follows a two-phase structure:
\begin{enumerate}
    \item \textbf{Inflow:} Multiple incoming wire transfers from the selected external sources, all in the user-defined timeframe, with a default of 24 hours.
    \item \textbf{Dispersal:} Following a small delay period, we generate cash withdrawal transactions. The individuals' cash holdings are modelled as a separate account (next to, for example, current and savings accounts). These withdrawals also happen with a varying interval in a specified time period. To account for hypothetical operational cost, the total outflow amount is typically 85-95\% of the total inflow amount.
\end{enumerate}

\subsubsection{Front business activity}
This pattern implements the use of front business behaviour described in Section~\ref{ref:network_topology}, where businesses deposit large amounts of cash, across accounts at different institutions, and immediately transfer the funds to overseas business accounts.

\begin{figure}[h!]
    \centering
    \includegraphics[width=0.5\linewidth]{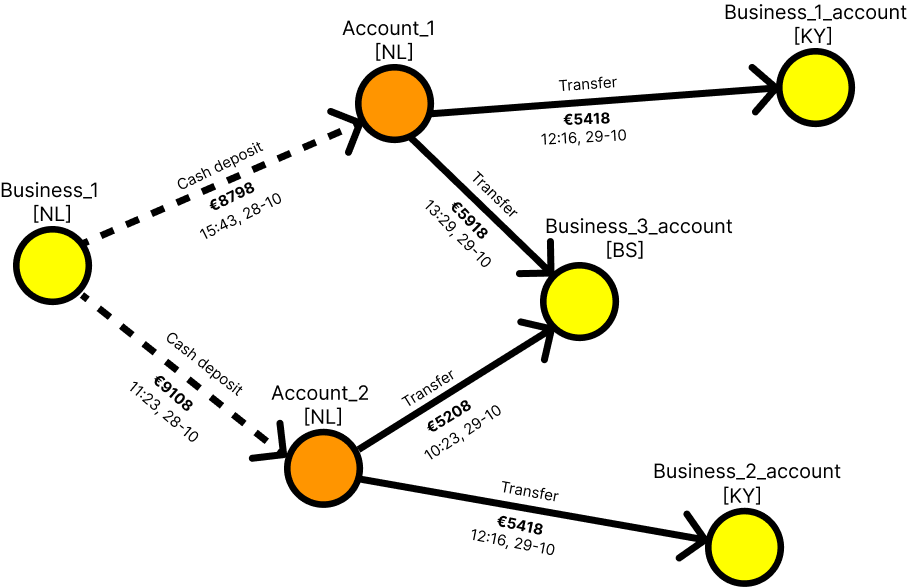}
    \caption{Visualisation of the front business pattern. \texttt{Business\_1} owns \texttt{Account\_1} and \texttt{Account\_2}.}
    \label{fig:frontbusiness}
\end{figure}

\paragraph{Structural component}
Tide identifies business entities from the high risk cluster that control multiple accounts across different banks. Additionally, it selects overseas business accounts where each account is in a different country and owned by another business entity.

\paragraph{Temporal component}
The implementation of this pattern consists of two phases:
\begin{enumerate}
    \item \textbf{Deposit phase:} Cash deposits occur across multiple accounts in a short timeframe. This cash comes from the global cash account in our financial system.
    \item \textbf{Transfer phase:} Following deposits with delays of 0.5 to 6 hours, transfers go to the overseas business accounts. The transfer amounts are 80-100\% of the total deposited amount, so transfers appear more as legitimate international business transfers.
\end{enumerate}

\subsubsection{Synchronised transactions}
This pattern implements the synchronised transaction behaviour described in the literature, where multiple seemingly unrelated entities perform coordinated transactions that suggest collusion.

\paragraph{Structural component}
Tide selects multiple coordinating entities with diverse characteristics to appear unrelated. For example, different age group, occupation, or different country. We also select a recipient entity, typically from high-risk jurisdiction clusters, that will receive the coordinated transfers.

\paragraph{Temporal component}
The implementation of this pattern generates the coordinated activity described in the literature within tight time windows:
\begin{enumerate}
    \item \textbf{Synchronised deposits:} Each coordinating entity receives cash deposits at nearly the same time. These deposits are distributed within an n-hour synchronisation window.
    \item \textbf{Aggregation:} After delays of 1-6 hours, coordinating entities transfer 85-95\% of deposited amounts to the recipient account. As noted in the literature, these transactions often involve amounts structured below reporting thresholds and similar transaction values.
\end{enumerate}

\subsubsection{U-Turn Transactions}
This pattern implements a round-trip transaction structure, where funds are transferred out through intermediaries before returning to accounts linked to the original sender.

\begin{figure}[h!]
    \centering
    \includegraphics[width=0.5\linewidth]{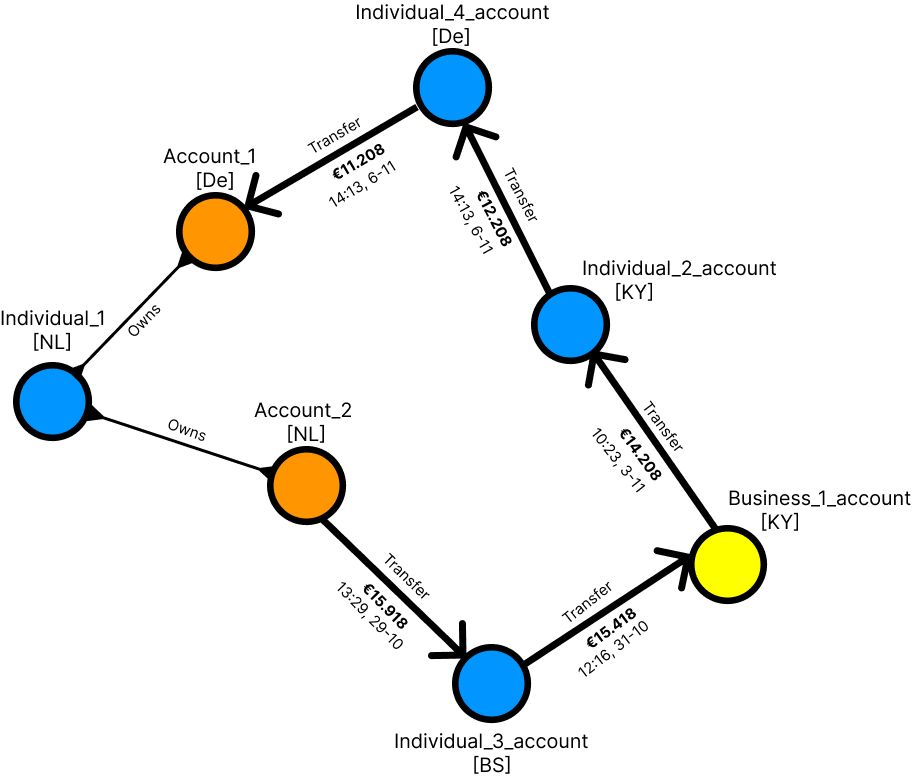}
    \caption{Visualisation of the U-turn transaction pattern.}
    \label{fig:enter-label}
\end{figure}

\paragraph{Structural component}
Tide selects a high-risk source entity with at least two accounts, intermediaries, and a destination account in a high-risk jurisdiction. Nodes are clustered as intermediaries based on some demographic and operational characteristics; our algorithm targets young adults, so people between 18–24 years old, who are often recruited as money mules due to financial vulnerability, and elderly people who are over 65 years old, who may be exploited. Next to age, we look at individuals with occupations that provide access to financial systems or are traditionally associated with high money laundering risks. For this, we look at the occupations that partake in cases of \cite{austrac2014typologies}, and extrapolate on that to other high paid occupations, or other occupations in the financial industry. Geographic location also influences selection, with preference given to individuals in high-risk jurisdictions where regulatory oversight may be weaker. For front business entities, the approach focuses on small companies with ten or fewer employees operating in a cash intensive business category.

\paragraph{Temporal component}
The implementation of this pattern constructs transaction paths from the source, to at least one account in a high-risk jurisdiction, through the set of intermediaries. Each transaction occurs with one to five day delays, with amounts decreasing by 1-3\% fees at each step.

\subsection{Simulating Legitimate Background Activity}
\label{sec:background_activity}
The simulation of background financial activity represents the normal, legitimate transactions that occur in a financial network. These transactions form the baseline behaviour against which suspicious patterns can be detected. We split different types of background activity into various patterns, where, like the fraudulent patterns, each pattern is composed of a structural and temporal aspect. We implement four types of base background activity: random everyday payments in a lower range, higher range payments, salary payments, and fraudster background activity. We also implement specific background patterns to counterbalance the very specific fraud patterns.

\subsubsection{Transaction Amount Calibration}
To prevent amount-based data leakage, all transaction amounts are drawn from
log-normal distributions calibrated to the 2022 Federal Reserve Payments Study
(FRPS, 8th Triennial)~\cite{frps2022}, the Bureau of Labor Statistics Usual
Weekly Earnings report~\cite{bls_weekly_earnings}, and ADP payroll
microdata~\cite{adp2025_payhistory}, following a similar methodology as
Altman et al.~\cite{altman2023realistic}. Many stochastic models in economics
and finance are described by distributions with a log-normal body and Pareto
tail~\cite{bee2022lognormal}; we approximate this using clipping to
type-specific $[\text{min},\text{max}]$ bounds. For each transaction type we
set $\mu = \ln M$ where $M$ is the target median, and select $\sigma$ large
enough to produce realistic heavy tails ($\sigma \geq 0.6$), accepting that
theoretical means may deviate from source averages.
Table~\ref{tab:amount-params} lists the base log-normal parameters alongside
the generated statistics. Structuring overlays and pattern-specific behaviour
shift the aggregate means above the theoretical values. Three types are
calibrated to FRPS averages (Payment, Transfer, Withdrawal); Deposit
(marked~$\dagger$) is a modelling assumption, as we were not able to find a public source that reports typical deposit amounts directly.

\begin{table}[h]
\centering\small
\caption{Log-normal amount parameters and generated statistics per transaction type in \textbf{\textit{LI}} dataset.}
\label{tab:amount-params}
\begin{tabular}{lrrrrrr}
\toprule
Type & $\mu$ & $\sigma$ & Source Avg & Gen.\ Median & Gen.\ Mean \\
\midrule
Payment    & 3.8 & 1.2 & \$60 (FRPS)         & \$46   & \$218    \\
Transfer   & 5.5 & 2.0 & \$1{,}634 (FRPS)    & \$221  & \$1{,}863 \\
Withdrawal & 4.8 & 0.9 & \$198 (FRPS)        & \$130  & \$546    \\
Deposit$^\dagger$ & 5.3 & 1.5 & ---           & \$220  & \$977    \\
\bottomrule
\end{tabular}
\end{table}

Structuring deposits use a uniform distribution over \$7,000--9,999, or its equivalent in other currencies, to model deliberate sub-threshold clustering; 3--5\% of legitimate background transactions also fall in this range to prevent structuring from becoming a pure fraud signal.

\subsubsection{Transaction Frequency Calibration}
Table~\ref{tab:frps-raw} reports noncash payment volumes from the 2022 Federal Reserve Payments Study~\cite{frps2022}. Card and ACH debit payments dominate U.S.\ noncash transactions ($\sim$91\%), while ACH credit transfers and ATM withdrawals account for $\sim$7\% and $\sim$2\%, respectively. However, our included fraud patterns primarily exploit transfers, deposits, and cash withdrawals rather than card purchases. We therefore over-represent these transaction types to ensure the generated data contains sufficient instances of the payment flows targeted by our money-laundering patterns (Table~\ref{tab:freq-params}). Note that FRPS tracks outgoing payments; deposits (incoming flows) are not reported. Salary payments are modelled as incoming transfers from employer accounts and subsumed under the Transfer category.

\begin{table}[htbp!]
\centering\small
\caption{Transaction type frequencies: FRPS reference vs.\ generated data in \textbf{\textit{LI}} dataset.}
\label{tab:freq-params}
\begin{tabular}{lrr}
\toprule
Type & FRPS Share$^\dagger$ & Generated Share \\
\midrule
Payment    & $\sim$91\% & 68\% \\
Transfer   & $\sim$7\%  & 12\% \\
Withdrawal & $\sim$2\%  & 8\%  \\
Deposit    & ---  & 11\% \\
\bottomrule
\end{tabular}
\begin{tablenotes}
\small
\item $^\dagger$ FRPS tracks outgoing payments only; deposits (incoming flows) are not measured. Our generated distribution includes deposits, which reduces the relative share of other types accordingly.
\end{tablenotes}
\end{table}

\subsubsection{Random everyday transactions}
This transaction category captures the baseline of legitimate background activity within the network. It encompasses low-value payments intended to simulate routine consumer purchases and recurring expenses, constrained within value ranges defined by the configuration parameters. The frequency of these interactions is governed by a user-specified rate, denoting the average daily transaction volume per account.

Like the fraud patterns, this pattern also consists of a structural component that selects legitimate accounts from pre-computed clusters, and a temporal component that generates transaction sequences over the specified time period. The structural component selects legit entities from our pre-computed cluster (As explained in Section~\ref{sub:clustering}).
The temporal component generates transactions using a memory-efficient streaming approach. The total number of expected transactions is calculated as:

\begin{equation}
\text{Total random transactions} = \lfloor r \times d \times |A| \rfloor
\end{equation}

Where $r$ is the transaction rate per account per day, $d$ is the total number of days in the simulation period, and where $|A|$ is the number of active accounts, including both legitimate and a subset of fraudulent accounts to prevent topological isolation. The count may be further capped by a configurable transaction budget.

Transaction types are selected probabilistically with configurable weights for wire transfers to other individuals, payments to businesses, deposits, and withdrawals. Transaction amounts are sampled from the log-normal distributions in Table~\ref{tab:amount-params}, with type-specific parameters for payments, transfers, deposits, and withdrawals.

\subsubsection{High-value transactions}
We define `high-value transactions' as legitimate transfers of significant magnitude that occur considerably less frequently than standard background activity.. These payments are meant to model transactions like luxury purchases, house purchases, and business transactions.

The pattern generates transactions between individuals and businesses. It identifies entities from two main sources: individuals with high-paid occupations and medium to large businesses (which we define as having more than 10 employees).

The total number of transactions to generate is calculated using the formula:

\begin{equation}
\text{Total high-value transactions} = \lfloor r_{monthly} \times M_{total} \times |A_{high}| \rfloor
\end{equation}

Where $r_{monthly}$ is the monthly transaction rate per high-value account (configured by the user), $M_{total}$ is the total number of months in the simulation period, and $|A_{high}|$ is the number of identified high-value accounts.

Transaction amounts are sampled from the high-value log-normal distribution (Table~\ref{tab:amount-params}) and rounded to the nearest \$100. All transactions are scheduled during business hours (9 AM to 5 PM) with random dates distributed throughout the simulation period.

\subsubsection{Salary payments}
The salary payments pattern models recurring compensation from businesses to individuals. The system designates legitimate business entities as payers and specific individuals as recipients. Each business is linked to 1--3 individual recipients; this simulates employer-employee relationships while preventing unrealistic density in the transaction graph.

The temporal component generates schedules based on standard payroll intervals: bi-weekly, monthly, or user-defined custom periods. For monthly cycles, payments align with specific calendar days (defaulting to the 1st, 15th, and 30th). Payment dates are calculated iteratively across the full simulation duration. Transaction amounts are sampled from the salary distribution (Table~\ref{tab:amount-params}), with minor random variations applied to simulate overtime or bonuses. All transactions occur during business hours, and the established employer-employee links remain static throughout the simulation.

\subsubsection{Fraudster background activity}
As established in Section~\ref{ref:temporal_patterns}, a common sign of fraudulent activity is a deviation from past transactional behaviour. To prevent classifiers from exploiting the absence of prior activity as a trivial fraud signal, entities that partake in fraudulent patterns also maintain a transactional history that is statistically indistinguishable from legitimate users.

The structural component selects fraudulent entities from our pre-computed cluster and pairs them with legitimate entities to serve as transaction counterparts (as explained in Section~\ref{sub:clustering}). Unlike earlier iterations of this system, the temporal component now generates transactions at the same rate as legitimate background activity, rather than at a reduced rate. The expected number of transactions for each fraudster account is calculated as:

\begin{equation}
\text{Fraudster transactions per account} = \max(1, \lfloor r \times d \rfloor)
\end{equation}

Where $r$ is the per-account daily transaction rate (identical to the rate used for legitimate random payments), and $d$ is the total number of days in the simulation period. The $\max$ function ensures each fraudster generates at least minimal activity.

Transaction amounts are drawn from the same log-normal distributions as legitimate payments, so that amount distributions do not distinguish fraudulent from legitimate accounts. Transaction direction is randomly determined with equal probability for incoming and outgoing payments. All transactions are executed during business hours.

\subsubsection{Counter-leakage background patterns}
\label{sec:counter_leakage}
A na\"ive approach to background activity generation can introduce data leakage: if certain transaction types, temporal signatures, or amount ranges appear exclusively in fraudulent patterns, classifiers can exploit these artefacts rather than learning genuine structural indicators of money laundering. For example, if cash deposits only occur in fraud patterns, then any deposit becomes a trivial fraud signal; similarly, if periodic transfers or multi-hop chains are exclusively fraudulent, temporal and topological features leak the label.

To address this, we introduce seven additional background patterns that inject legitimate activity overlapping with fraud signatures along three dimensions.

\paragraph{Temporal overlap}
Legitimate bursts generate tight clusters of 5--15 transactions within minutes, modelling shopping sprees or bill-payment runs, so that high-frequency activity is no longer exclusive to fraud.

\paragraph{Structural overlap}
Legitimate transaction chains create multi-hop paths ($A \rightarrow B \rightarrow C \rightarrow D$) of length 2--5, modelling group expenses, family support flows, and business supply chains. These are topologically similar to layering hops but are labelled as legitimate.
Legitimate rapid fund flows generate inflow--outflow sequences resembling the rapid fund movement pattern, based on scenarios such as account consolidation, emergency expenses, or event funding.

\paragraph{Feature overlap}
Legitimate cash operations inject deposits and withdrawals sampled from the calibrated log-normal distributions (Table~\ref{tab:amount-params}), so that cash transaction types are not exclusive to fraud. A rapid-deposit mode generates 3--5 deposits within hours, mimicking small-business cash handling.
Legitimate structuring generates deposits in the \$7,500--\$9,999 range for legitimate reasons (house savings, freelance income, small-business cash handling), with an optional burst timing mode that clusters deposits within hours, matching the temporal signature of fraudulent structuring. Unlike other transaction types, structuring amounts remain uniformly distributed in this sub-threshold range, as this deliberate clustering is the behavioral signal AML systems must detect.

Legitimate high-risk entity activity ensures that entities with high risk scores, from high-risk jurisdictions, or in high-risk business categories also engage in normal transaction activity, preventing the risk score from being a near-perfect fraud predictor.
All counter-leakage patterns also include fraudulent accounts among their participants (typically 50--90\% of fraud accounts are mixed in), ensuring that fraudulent entities are not topologically isolated from legitimate activity.

Beyond structural and temporal overlap, all counter-leakage patterns sample transaction amounts from the same log-normal distributions as standard background activity, preventing amount-based leakage.
Table~\ref{tab:background_overview} summarises all background patterns.

\subsection{Framework Configurations and Additional Features}
The data generation pipeline manages configuration, feature processing, and data export through a unified workflow.

\subsubsection{Parameterization and Reproducibility}The system is controlled by two YAML configuration files. The first file manages general graph-level parameters, such as the total number of entities and the simulation timeframe. The second file defines pattern-specific parameters that specify the structural and transactional properties of the money-laundering topologies. To ensure reproducibility, a random seed is applied at every step of the pipeline, making the graph structure, attribute assignment, and transaction sequence fully deterministic.

\subsubsection{Feature Selection and Data Splitting}To simulate a blind detection scenario, internal attributes used during generation (e.g., \texttt{risk\_score}, \texttt{occupation}, \texttt{business\_category}) are excluded from the final dataset. The exported feature matrix contains only raw transactional fields, such as timestamp, amount, and currency. The \texttt{is\_fraudulent} label is retained as a target variable but excluded from the node features. For a detailed overview, see Table~\ref{tab:exported_features}.

\subsubsection{Data Export}
Output formats are configured via boolean flags in the configuration YAML. The system supports serialization into raw CSV files (node and edge lists) and direct export to graph objects for Python-based frameworks, specifically PyTorch Geometric (\texttt{.pt}) and NetworkX (\texttt{.gpickle}). Additionally, JSON metadata is generated for each pattern instance, recording the participating entities and transaction sequences.

\subsubsection{Availability}
The Tide data generation framework is open-source and publicly available at \url{https://github.com/mntijn/Tide}.

\subsection{Reference Benchmark: Datasets and Pipeline}
\label{sec:reference_dataset}

\paragraph{Dataset Statistics}
We publish two dataset variants generated from the same base population of 8{,}000 individuals, differing in their target illicit ratio. Both datasets simulate a 12-month period (January--December 2025) and share the same random seed, graph topology, and background activity configuration; only the fraud injection parameters differ. Table~\ref{tab:dataset_variants} summarises the two variants.

\begin{table}[t]
\centering
\caption{Published dataset variants. Both share the same base graph (8{,}000 individuals, 12-month simulation).}
\label{tab:dataset_variants}
\begin{tabular}{@{}lrrrr@{}}
\toprule
\textbf{Variant} & \textbf{Nodes} & \textbf{Edges} & \textbf{Illicit ratio} & \textbf{Imbalance} \\
\midrule
\textsf{\textbf{LI}} & 36{,}629 & 7{,}642{,}030 & 0.10\% & 959:1 \\
\textsf{\textbf{HI}} & 36{,}653 & 7{,}618{,}299 & 0.19\% & 531:1 \\
\bottomrule
\end{tabular}
\end{table}

The low-illicit variant (\textsf{LI}) and high-illicit variant (\textsf{HI}) use identical layering depth (2--5 hops) and fraud selection parameters, differing only in the number of injected fraud patterns (90 vs.\ 320). The graph contains three node types: accounts, individuals, and businesses.

Background transaction volume is derived from the target fraud ratio: the system generates fraud patterns first, then scales legitimate background activity such that $\text{fraud edges} / \text{total edges} \approx \text{target ratio}$. The per-account daily transaction rate is capped at 2 and scaled down as needed to meet the target ratio.

\paragraph{Data Partitioning}
To prevent temporal data leakage and to simulate operational deployment scenarios, we adopt the temporal split strategy proposed by Altman et al. \cite{altman2023realistic}. Transactions are ordered by timestamp and partitioned using two temporal thresholds, $t_1$ and $t_2$. Training indices correspond to transactions before $t_1$, validation indices to transactions in the interval $[t_1, t_2)$, and test indices to transactions after $t_2$. The split ratios are 60\%, 20\%, and 20\% of the chronologically ordered transaction sequence, respectively.

For GNN models, we construct three graphs to reflect the dynamic nature of financial transaction networks. The training graph $G_{\text{train}}$ contains only transactions before $t_1$. The validation graph $G_{\text{val}}$ contains all transactions before $t_2$ (i.e., training and validation transactions), but evaluation is performed exclusively on validation indices. The test graph $G_{\text{test}}$ contains all transactions up to $t_{\text{max}}$, with evaluation restricted to test indices. This construction simulates the operational scenario where models must classify new transaction batches while having access to historical transaction graphs for pattern identification. This methodology ensures that models are evaluated on their ability to detect future fraud given a complete historical context, reflecting the temporal constraints of production AML systems.

For GBT models (LightGBM and XGBoost), we apply the same temporal partitioning strategy but convert the graph data to tabular feature vectors. Each transaction is represented by a feature vector comprising edge attributes (amount, timestamp, currency), source-node features, and destination-node features. The training set uses transactions from $G_{\text{train}}$, while validation predictions are computed only for new transactions in the interval $[t_1, t_2)$, despite the feature extractor having access to $G_{\text{val}}$ for computing graph-based features. This ensures that GBT models are evaluated under the same temporal constraints as GNN models, thereby maintaining comparability across model classes.

\paragraph{Pipeline Implementation}
We implement and publish the experimental workflow as a modular Kedro pipeline to ensure reproducibility. The pipeline handles data loading, temporal splitting, model training (GNN and GBT), evaluation, and result serialization. Hyperparameters are externalized, allowing researchers to reproduce our benchmarks or evaluate new models against the published datasets with minimal modification.

\paragraph{Availability}
Both dataset variants (\textsf{LI} and \textsf{HI}) are publicly available on Zenodo at \url{https://zenodo.org/records/18804069}. The Kedro pipeline and all accompanying code are released at \url{https://github.com/mntijn/tide-model-comparison}.


\section{Evaluation}
\label{sec:evaluation}

We split the evaluation into three parts: we first validate the correctness of generated patterns, we then assess the system's scalability and performance, and finally, we evaluate the quality of the generated datasets through the performance of supervised machine learning models trained on transaction classification tasks. The experimental design follows the methodology described in Section~\ref{sec:methodology}, employing both GNN and GBT baselines under identical temporal split configurations.

\subsection{Structural Validation}
\label{sec:pattern_effectiveness}

The structural validation assesses how well Tide generates money-laundering patterns that conform to their formal definitions. We analyse the structural accuracy, financial compliance, and temporal correctness of generated patterns across all three pattern types. The results demonstrate whether our pattern generation algorithms successfully implement the theoretical frameworks established in Section~\ref{sec:related}.

Additionally, the pattern visualisation results provide insights into the quality and recognisability of the generated patterns, indicating whether they exhibit the characteristic structures expected in real money laundering scenarios.

For each pattern, we show our test results and a visualisation of both the structure and the timeline of transactions.

\subsubsection{Front Business Pattern}
Our validation confirmed that all generated instances of the Front Business Activity pattern met their defined constraints. The tests verified a structure where 5 to 15 large cash deposits between €15,000 and €75,000 are each rapidly layered through overseas transfers. Each transfer, amounting to 80-100\% of the deposit, was executed within 0.5 to 6 hours of the deposit, which is consistent across all tested instances.

\begin{figure}[htbp]
    \centering
    \begin{subfigure}[t]{0.48\textwidth}
        \centering
        \includegraphics[width=\linewidth]{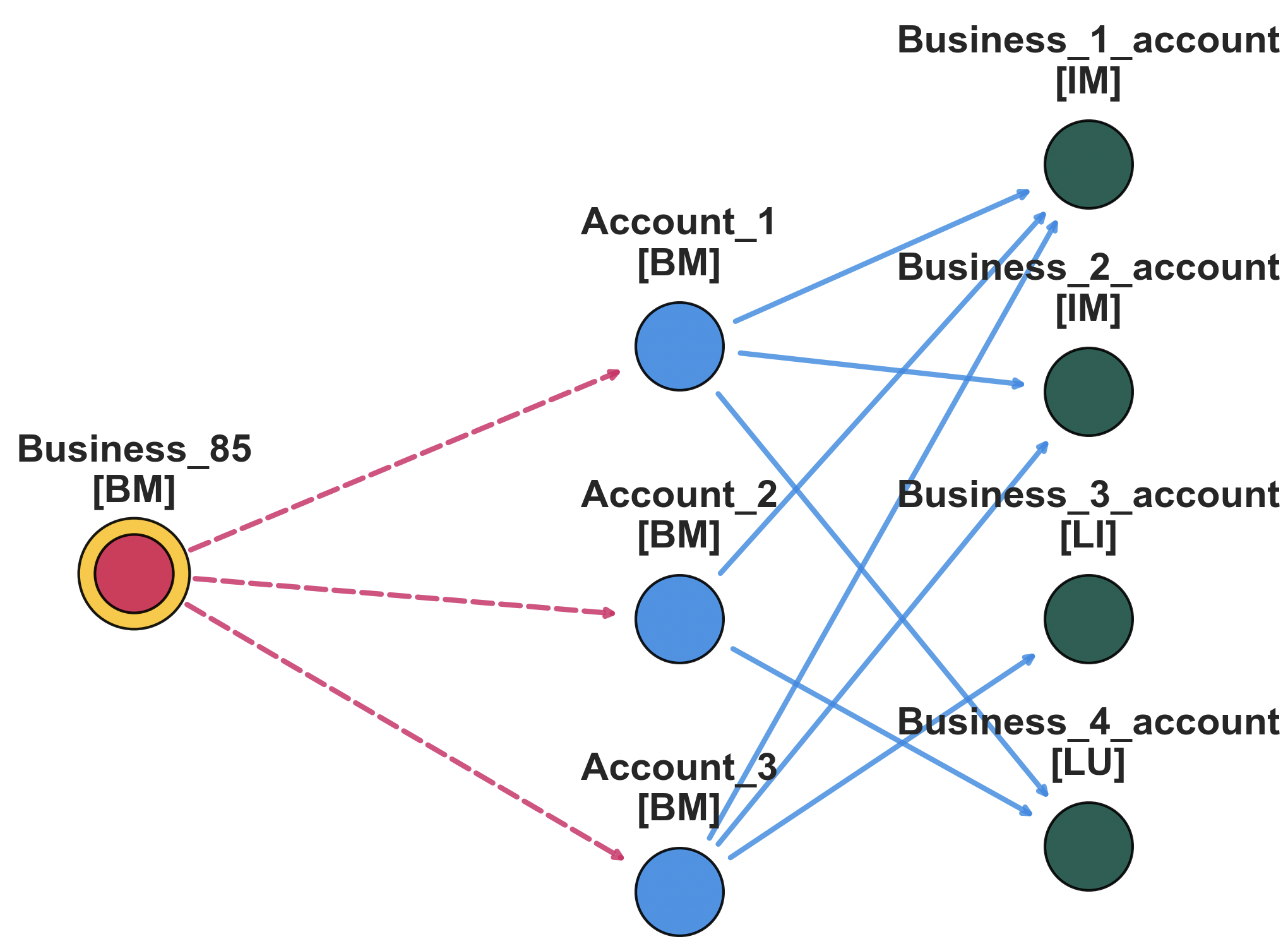}
        \caption{The front business activity pattern. The source business (red-yellow node on the left) deposits funds into its accounts (blue nodes), which is then followed by transfers to overseas businesses (green nodes on the right). Brackets display the country that the node is located in.}
        \label{fig:front_business_network}
    \end{subfigure}
    \hfill
    \begin{subfigure}[t]{0.48\textwidth}
        \centering
        \includegraphics[width=\linewidth]{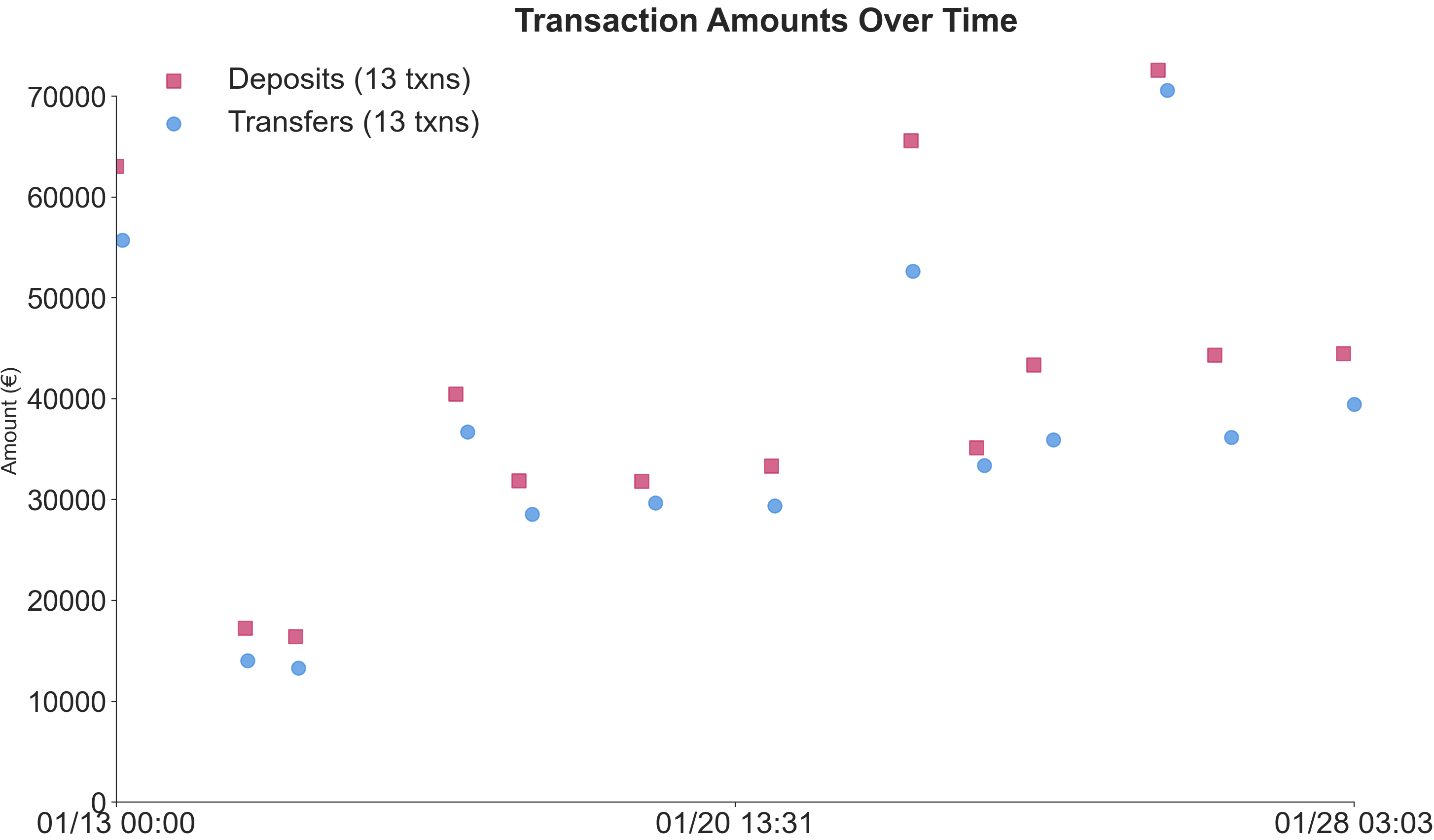}
        \caption{Timeline of the transactions. All transfers happen within hours after deposits.}
        \label{fig:front_business_timeline}
    \end{subfigure}
    \caption{The front business pattern shown as a network graph and a transaction timeline.}
    \label{fig:front_business}
\end{figure}

The isolated generated pattern topology and the corresponding timeline of deposits and transfers are shown in Figure~\ref{fig:front_business}. In Figure~\ref{fig:front_business_network}, on the left one can see the source business, that deposits funds into its accounts, which is then followed by transfers to other overseas businesses. This is visible on the timeline (Figure~\ref{fig:front_business_timeline}), where all transfers follow quickly after the deposits.

\subsubsection{Rapid Fund Movement Pattern}
Our tests showed that all three generated instances successfully met the predefined constraints.

We examined the attributes as we detailed before. The financial and timing constraints were all satisfied. All individual inflow transactions stayed below the €10,000 reporting threshold. Second, we calculated the outflow-to-inflow ratio for each instance and found that all three fell within the required 0.85-0.95 range. Third, we verified the timing requirements and found that the transaction ordering was correct in all cases. The inflow phase completed within 24 hours, there was at least a 1-hour delay before withdrawals began, and the entire pattern finished well under the 128-hour maximum duration.

\begin{figure}[htbp]
    \centering
    \begin{subfigure}[b]{0.48\textwidth}
        \centering
        \includegraphics[width=\linewidth]{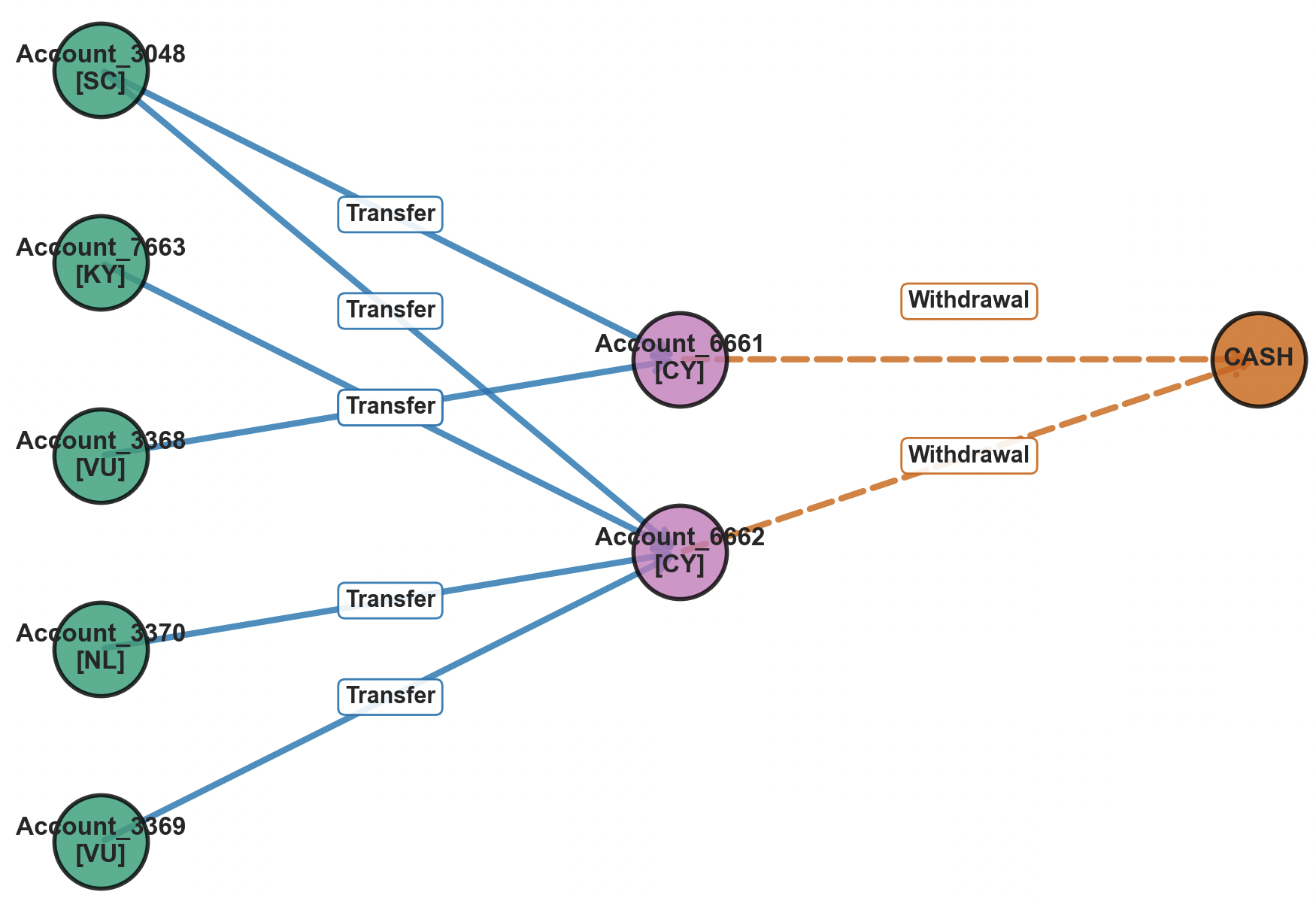}
        \caption{The rapid fund movement pattern. Overseas accounts (green nodes) first transfer money into multiple accounts of the same entity (purple nodes), which is then withdrawn. Brackets display the country that the node is located in.}
        \label{fig:rapid_fund_movement_network}
    \end{subfigure}
    \hfill
    \begin{subfigure}[b]{0.48\textwidth}
        \centering
        \includegraphics[width=\linewidth]{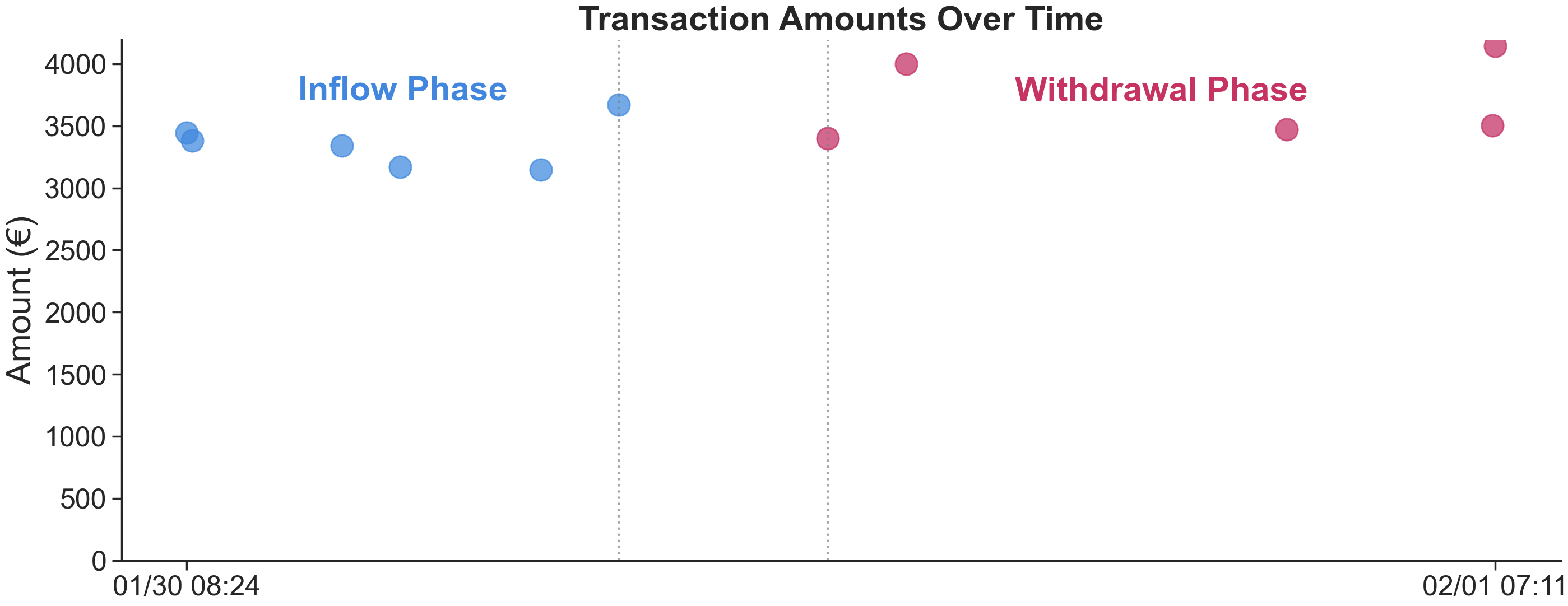}
        \caption{Timeline of the transactions over simulation time. First there is a `burst' of incoming transfers (blue) into the accounts, which is later followed by a burst of withdrawals (pink).}
        \label{fig:rapid_fund_movement_timeline}
    \end{subfigure}
    \caption{The rapid fund movement pattern shown as a network graph and a transaction timeline.}
    \label{fig:rapid_fund_movement}
\end{figure}

The topology in Figure~\ref{fig:rapid_fund_movement_network} shows a similar structure as illustrated in Section~\ref{meth:rapidmovement}. The overseas accounts first transfer money to multiple accounts owned by the beneficiary, after which these funds are withdrawn. Additionally, the timeline in Figure~\ref{fig:rapid_fund_movement_timeline} demonstrates the rapid succession of inflows within a compressed timeframe, followed by the distinct outflow phase.

\subsubsection{Repeated Overseas Transfer Pattern}
Our tests demonstrated that all three generated instances successfully satisfied the predefined constraints for the Repeated Overseas Transfer pattern.

We analysed the pattern attributes using the same approach applied to previous patterns. The transfer frequency requirements were met across all instances, with each pattern generating between 4-12 individual transfers as specified. We verified that every transfer amount fell within the required €5,000-€20,000 range. The destination diversity constraint was also satisfied, with each instance directing transfers to 2-5 distinct overseas accounts as intended. All patterns maintained regular scheduling at 7-, 14-, or 30-day intervals throughout the sequence. The temporal consistency remained stable across the entire pattern duration, with no significant deviations from the established intervals. Additionally, we validated that all destination accounts were properly classified as overseas entities.

\begin{figure}[htbp]
    \centering
    \begin{subfigure}[b]{\textwidth}
        \centering
        \includegraphics[width=0.8\linewidth]{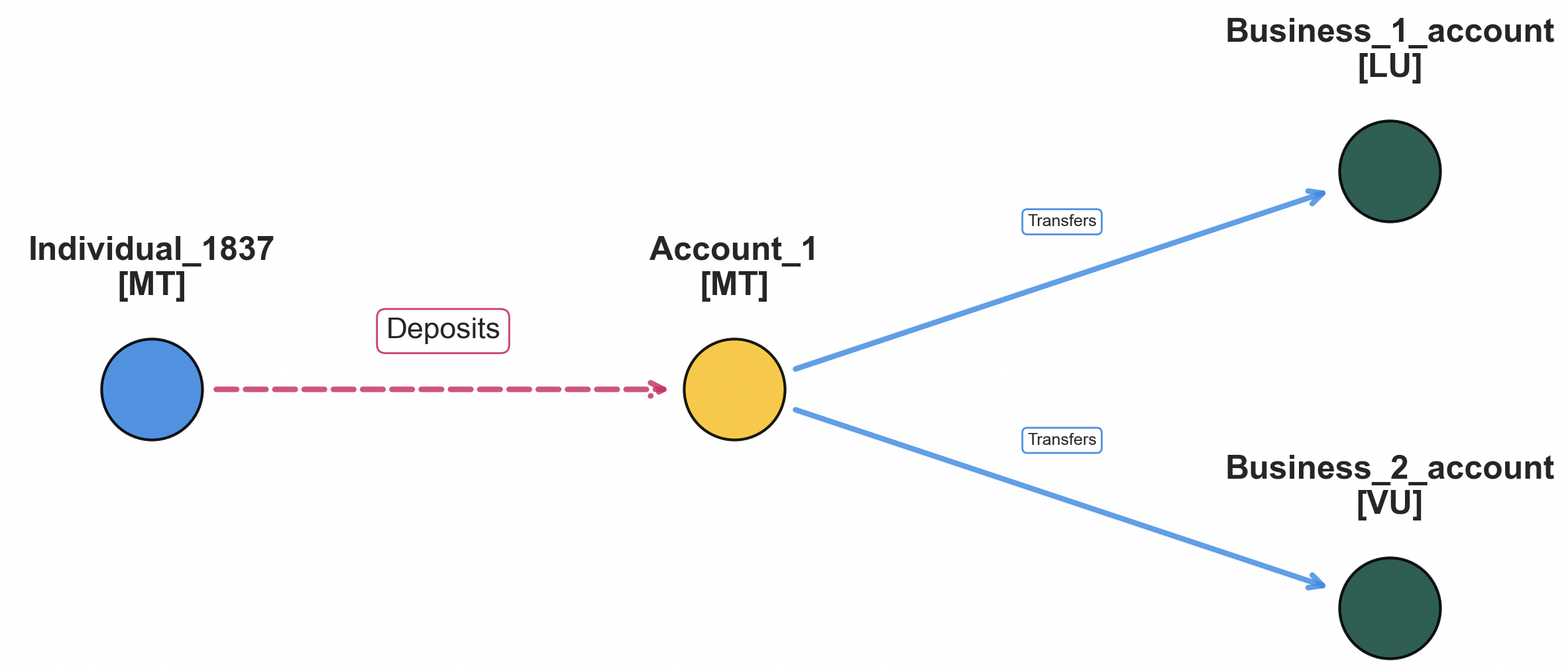}
        \caption{The repeated overseas transfers pattern. An individual (blue node on the left) repeatedly deposits money into their account (middle yellow node), which is then followed by repeated overseas transfers to the same beneficiaries (green nodes on the right). Brackets display the country that the node is located in.}
        \label{fig:repeated_transfers_network}
    \end{subfigure}
    
    \vspace{1em}

    \begin{subfigure}[t]{0.48\textwidth}
        \centering
        \includegraphics[width=\linewidth]{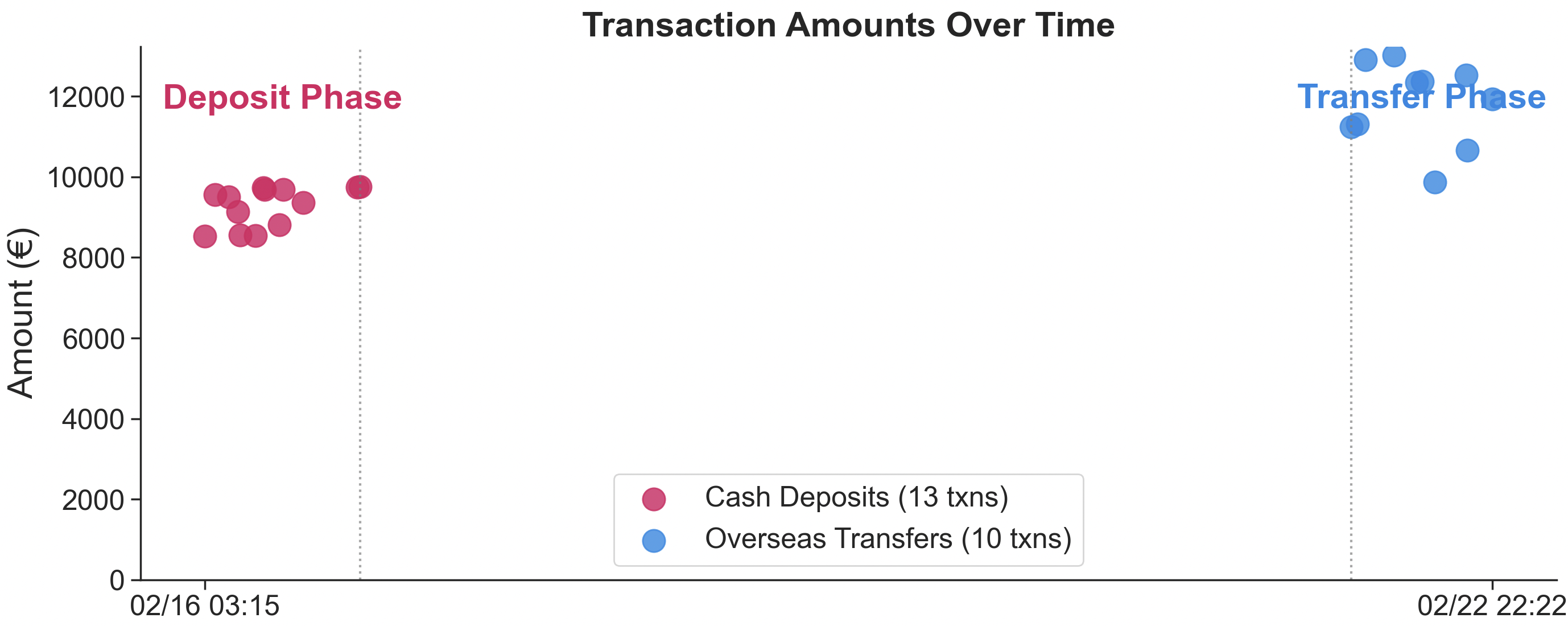}
        \caption{Timeline of the high-frequency transactions over simulation time. The transfer phase consists of a similar burst of transactions as the deposit phase.}
        \label{fig:repeated_transfers_timeline}
    \end{subfigure}
    \hfill
    \begin{subfigure}[t]{0.48\textwidth}
        \centering
        \includegraphics[width=\linewidth]{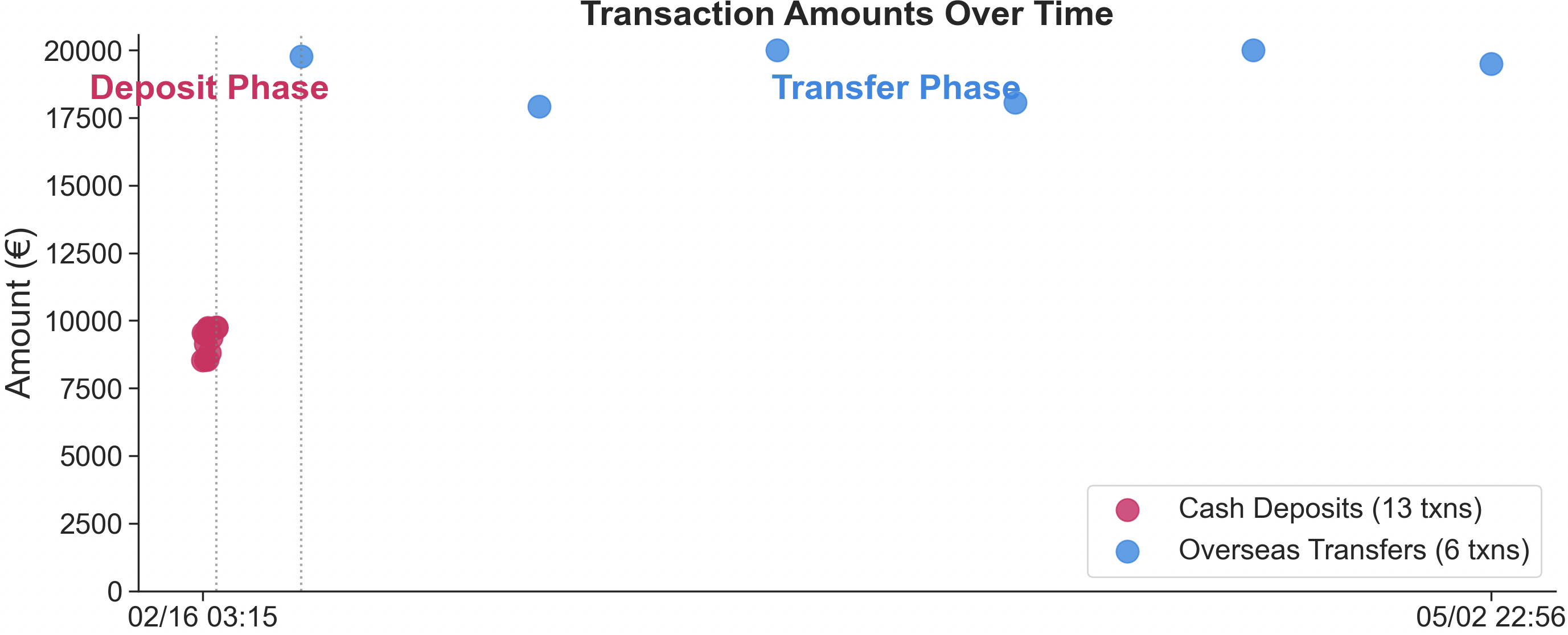}
        \caption{Timeline of the periodic transactions. The deposit phase consists of frequent deposits followed by transfers every $\approx 7$ days.}
        \label{fig:repeated_transfers_timeline_2}
    \end{subfigure}
    
    \caption{The repeated overseas transfer pattern shown as a network graph and two distinct transaction timelines: one for high-frequency activity and one for periodic behaviour.}
    \label{fig:repeated_transfers}
\end{figure}

The isolated generated pattern topology and the corresponding timeline of deposits and transfers are shown in Figure~\ref{fig:repeated_transfers}.

\subsubsection{U-turn transactions}
Our tests showed that all three generated instances successfully met the predefined constraints for the U-Turn transactions pattern. We examined the attributes following the same approach we used before. The chain length requirements were satisfied in all cases, with each instance containing between 4-7 entities as specified. Additionally, we checked that every pattern started with an international transfer in the correct €10,000-€100,000 range. We also calculated the fees applied at each intermediary hop and found that all instances properly applied fees between 1-3\% of the transaction amount at every step. We found that each pattern successfully returned between 70-90\% of the remaining amount back to the source entity after accounting for cumulative fees.

We also checked the timing constraints and found that all inter-hop delays stayed within the required 1-5 day range. The transaction sequences maintained proper ordering throughout each chain, and the complete U-Turn patterns finished within reasonable timeframes while meeting all specified timing requirements.

\begin{figure}[htbp]
    \centering
    \begin{subfigure}[b]{0.57\textwidth}
        \centering
        \includegraphics[width=\linewidth]{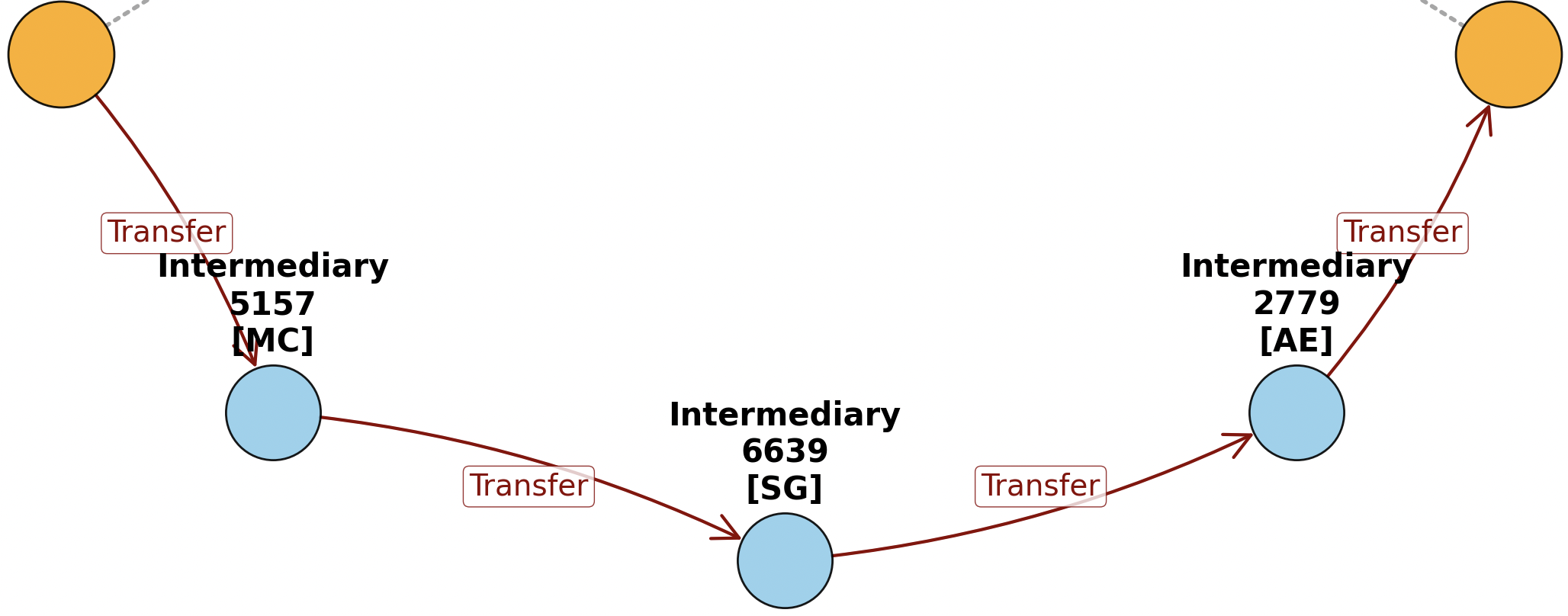}
        \caption{U-turn transaction pattern. The orange nodes are accounts, which controlled by the same entity. Funds flow from the left to the right node. Brackets display the country that the node is located in.}
        \label{fig:u_turn_network}
    \end{subfigure}
    \hfill
    \begin{subfigure}[b]{0.75\textwidth}
        \centering
        \includegraphics[width=\linewidth]{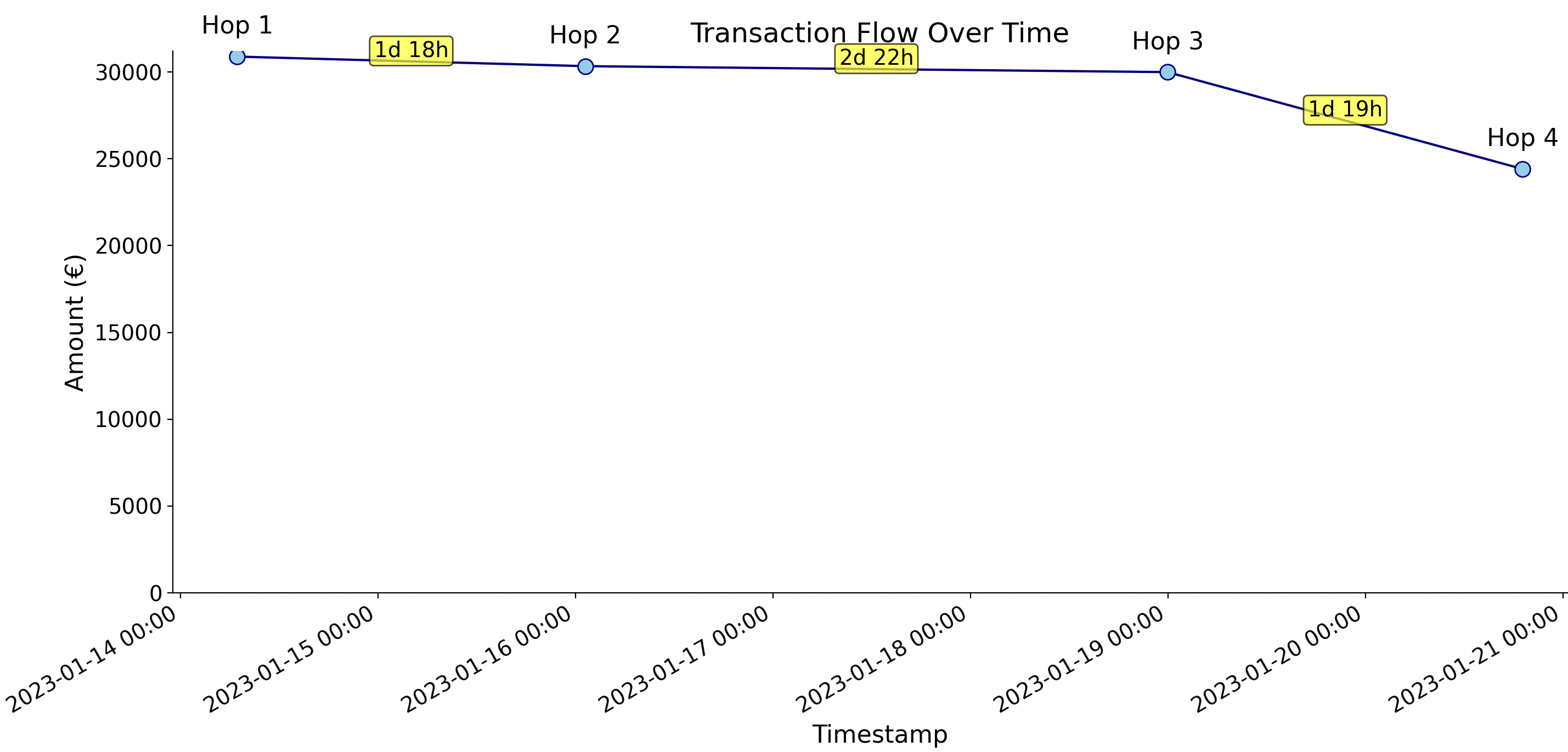}
        \caption{Timeline of the U-turn transaction pattern.}
        \label{fig:u_turn_timeline}
    \end{subfigure}
    \caption{The U-turn transaction pattern shown as a network graph and a transaction timeline.}
    \label{fig:u_turn}
\end{figure}

The isolated generated pattern topology and its timeline are shown in Figure~\ref{fig:u_turn}. Both orange accounts are owned by the same entity, either directly or indirectly through an owned business.

\subsection{Machine Learning Model Performance Evaluation}

\subsubsection{Machine Learning Models}
\label{sec:ml_models}

To evaluate the quality of the datasets generated by \textit{Tide}, we employ a spectrum of supervised learning models, with both Graph Neural Networks (GNNs) and Gradient-Boosted Trees (GBTs). Our selection criterion is both based on Altman et al. \cite{altman2023realistic}, and diagnostic: each model architecture is sensitive to a different modality of the generated data. By observing which models succeed or fail, we can disentangle the contributions of topological structure, transactional attributes, and temporal dynamics within our synthetic patterns.

\paragraph{GIN (Graph Isomorphism Network)}
We select GIN \cite{xu2019powerful_gnn} as our primary baseline for structural validation. GIN is theoretically designed to be as powerful as the Weisfeiler-Lehman (WL) isomorphism test, making it the ideal candidate to determine if the topological shapes of our injected fraud patterns are distinguishable from background activity.
\begin{itemize}
    \item \textbf{Mechanism:} Unlike standard GCNs that average neighbor features, GIN utilizes a sum aggregator. The node update rule is defined as:
    \begin{equation}
        h_v^{(k)} = \text{MLP}^{(k)} \left( (1 + \epsilon^{(k)}) \cdot h_v^{(k-1)} + \sum_{u \in \mathcal{N}(v)} h_u^{(k-1)} \right)
    \end{equation}
    \item \textbf{relevance to Dataset:} The sum aggregator allows GIN to distinguish neighborhoods based on cardinality and total volume rather than just distribution. In the context of our generated data, this tests whether the specific connectivity patterns of money laundering are structurally preserved and learnable.
\end{itemize}

\paragraph{GIN with Edge Updates (GIN+EU)}
While standard GIN excels at structure, it treats edges as binary connections. To evaluate the fidelity of the transactional attributes generated by Tide (specifically amounts and timestamps), we employ a GIN variant extended with edge updates. We use the exact implementation by Altman et al. \cite{altman2023realistic}.
\begin{itemize}
    \item \textbf{Mechanism:} This architecture explicitly incorporates edge features $e_{uv}$ into the message-passing framework. The message from neighbor $u$ to node $v$ transforms from a simple node feature $h_u$ to a fused representation $\phi(h_u, e_{uv})$.
    \item \textbf{Relevance to Dataset:} This model tests the "Interaction Hypothesis": does the combination of \textit{who} is transacting and \textit{how much/when} they transact provide a stronger signal than structure alone? High performance here would confirm that Tide's edge attributes contain discriminative signal beyond the graph topology.
\end{itemize}

\paragraph{PNA (Principal Neighbourhood Aggregation)}
Real-world financial crime often manifests as subtle statistical anomalies rather than obvious structural shapes. We include PNA \cite{corso2020pna} to test the distributional complexity of the generated data.
\begin{itemize}
    \item \textbf{Mechanism:} PNA generalizes the message passing by employing multiple aggregators (Mean, Max, Min, Std) and degree-scalers simultaneously.
    \item \textbf{Relevance to Dataset:} If our generator successfully creates complex laundering typologies (e.g., structuring amounts to minimize variance, or high-value outliers), PNA is the architecture best equipped to detect these second-order statistical properties.
\end{itemize}

\paragraph{Gradient Boosted Trees (Tabular Baselines)}
Finally, we evaluate LightGBM and XGBoost to isolate the quality of the temporal dynamics independent of graph topology. For this reason, even though our model selection is based on Altman et al. \cite{altman2023realistic}, we do not employ the same Graph Feature Preprocessor for these tabular models.
\begin{itemize}
    \item \textbf{Mechanism:} These models operate on "flattened" tabular data. Unlike GNNs, they cannot see the network structure directly. Instead, they rely on the manually engineered temporal features described in the preprocessing step (very much learning from features like time since last transaction).
    \item \textbf{Relevance to Dataset:} These baselines provide a crucial control. If GBTs perform well, it validates that \textit{Tide} generates temporally consistent sequences that are predictive in their own right. It proves that the temporal signal (bursts, delays, periodicities) is strong enough to drive detection even when the explicit graph structure is removed.
\end{itemize}

\subsubsection{Performance Assessment Metrics}

All metrics are computed on the test set using predictions from the best model checkpoint, selected based on the minimum validation loss. As detailed in Section~\ref{sec:performance_metrics}, we report standard classification metrics including F1 Score, Precision, Recall, and Area Under the Precision-Recall Curve (PR-AUC).

For all models, we perform threshold optimization on the validation set to determine the decision boundary that maximizes Youden's J statistic. This optimal threshold is then applied to the test predictions. This approach simulates operational deployment, where the system is calibrated on historical data to find the cut-off point that best separates illicit and legitimate transactions. In contrast to F1 optimization, which balances precision and recall, maximizing Youden's J identifies the threshold with the greatest statistical separation between the positive and negative class distributions. This ensures the decision boundary is driven by the model's fundamental discriminatory power rather than the specific prevalence of fraud in the validation set.

\paragraph{Temporal Feature Preprocessing}
Prior to GNN training, the temporal split procedure augments static node features with eight temporal statistics: in-degree, out-degree, and transaction amount statistics (mean, standard deviation, maximum) for both incoming and outgoing edges. These features are computed exclusively from edges available up to each split's temporal boundary: training features use only training edges, validation features use training edges, and test features use training and validation edges. This prevents temporal leakage by ensuring that node features never aggregate information from future transactions. Consequently, GNN models receive rich temporal features as input, reducing the representational burden on the message-passing layers.

\subsubsection{Experimental Results}
\label{sec:results}

Table~\ref{tab:combined_results} summarises PR-AUC performance across two experimental conditions: LI (0.10\% fraud rate) and HI (0.19\% fraud rate). Full results including F1, Precision, Recall, and Lift are provided in Appendix Tables~\ref{tab:results_li}--\ref{tab:results_hi}. PR-AUC serves as our primary evaluation metric given its robustness to class imbalance; threshold-dependent metrics are computed at Youden's optimal threshold for completeness. Results report mean $\pm$ standard deviation over five runs.

\paragraph{Results}
In the LI condition (0.10\%), LightGBM achieves the highest PR-AUC ($78.05 \pm 3.61$) and Lift ($749.19\times$), followed by XGBoost ($72.01 \pm 0.50$) and GIN ($69.86 \pm 6.55$). At higher illicit ratio (HI, 0.19\%), XGBoost achieves the strongest PR-AUC ($85.12 \pm 0.52$) and Lift ($452.78\times$). GIN shows stable PR-AUC across conditions ($69.86 \to 71.14$), while PNA improves from $69.20$ to $75.63 \pm 5.85$.

Precision remains low across most models (2--12\%) despite strong PR-AUC and Youden's $J$ scores. This is expected: Youden's threshold optimises sensitivity and specificity rather than precision, and under extreme class imbalance, even a modest false positive rate produces many false alarms relative to the small number of true positives. In operational AML systems, threshold selection is typically guided by alert-handling capacity and cost constraints rather than statistical optimisation criteria such as Youden's $J$; we report these metrics for methodological comparability rather than as deployment recommendations.

The +GFP variants consistently underperform their base models, likely because the generic graph features extracted by the preprocessor do not align with the specific structural and temporal patterns encoded by Tide's fraud injection mechanism. GIN+EU exhibits high variance across runs (e.g., $39.43 \pm 20.99$ in LI), suggesting sensitivity to initialisation that may limit its reliability in practice.

ROC-AUC exceeds $93\%$ for all models across both conditions, confirming strong class separability but illustrating why ROC-AUC is inadequate for AML evaluation: the metric fails to capture the precision challenges inherent to extreme class imbalance.

\paragraph{Analysis}
These results validate Tide as a benchmarking tool for AML detection research. The generator produces realistic class imbalance scenarios that differentiate model robustness: tabular models (XGBoost, LightGBM) maintain or improve PR-AUC as fraud prevalence increases, while GNN variants exhibit higher variance. The divergent model rankings across conditions (LightGBM optimal at 0.10\%; XGBoost optimal at 0.19\%) demonstrate that Tide's reference datasets can meaningfully distinguish model capabilities, supporting their use as standardised benchmarks for future work.

\begin{table}[htbp]
\centering
\caption{PR-AUC across experimental conditions. Bold indicates best performance per condition. Results are mean $\pm$ standard deviation over 5 runs.}
\label{tab:combined_results}
\begin{tabular}{@{}lrr@{}}
\toprule
\textbf{Model} & \multicolumn{2}{c}{\textbf{Illicit Ratio}} \\
\cmidrule(lr){2-3}
 & \textbf{LI (0.10\%)} & \textbf{HI (0.19\%)} \\
\midrule
GIN          & $69.86 \pm 6.55$  & $71.14 \pm 8.70$ \\
GIN+EU       & $39.43 \pm 20.99$ & $48.42 \pm 17.05$ \\
PNA          & $69.20 \pm 0.00$  & $75.63 \pm 5.85$ \\
LightGBM+GFP & $21.45 \pm 13.23$ & $64.40 \pm 24.36$ \\
XGBoost+GFP  & $35.45 \pm 2.59$  & $32.20 \pm 1.68$ \\
\midrule
LightGBM     & $\mathbf{78.05 \pm 3.61}$  & $70.47 \pm 27.91$ \\
XGBoost      & $72.01 \pm 0.50$  & $\mathbf{85.12 \pm 0.52}$ \\
\bottomrule
\end{tabular}
\end{table}

\section{Conclusion}
\label{sec:conclusion}
We presented Tide, a synthetic dataset generator that addresses the scarcity of accessible financial transaction data for Anti-Money Laundering research. Unlike existing generators that model network topology while neglecting temporal dynamics, Tide explicitly captures the interdependence between structural patterns and temporal characteristics, enabling injection of complex typologies including high-frequency bursts, synchronised transactions, and U-turn flows. The system produces deterministic, reproducible outputs through seed-based generation, and we release both the generator and reference datasets as open-source resources for the research community.

Our evaluation across two dataset variants with differing illicit ratios (LI: 0.10\%, HI: 0.19\%) revealed condition-dependent model rankings. LightGBM achieved the highest PR-AUC ($78.05$) in the low illicit ratio condition, while XGBoost performed best ($85.12$) at higher fraud prevalence. The +GFP variants consistently underperformed their base models, likely because generic graph features do not align with Tide's specific fraud patterns. These divergent rankings demonstrate that the reference datasets can meaningfully differentiate model capabilities across operational conditions.

\paragraph{Limitations}
The current work has several limitations that should be acknowledged. First, we did not disaggregate performance by pattern type, so it remains unclear whether certain typologies (e.g., U-turn vs. rapid fund movement) favour different architectures or present varying levels of detection difficulty.

Second, while Tide addresses key gaps in existing generators, the implemented pattern library covers only five laundering typologies. Real-world money laundering encompasses a broader, and more complex spectrum of schemes, including trade-based laundering, real estate manipulation, and cryptocurrency mixing, and more, which are not yet modelled.

Third, our transaction amount and frequency calibrations are derived primarily from U.S. sources (Federal Reserve Payments Study, Bureau of Labor Statistics), which may limit generalisability to financial systems with substantially different transaction profiles. Similarly, the risk scoring weights, while grounded in regulatory literature, represent simplified approximations of the complex factors that determine actual laundering risk.

Fourth, the absence of ground-truth validation against real financial data (an inherent constraint given privacy and legal barriers) means we cannot directly verify that the synthetic patterns reflect the statistical signatures of actual laundering schemes. The evaluation relies on the assumption that patterns derived from regulatory typologies are representative.

\paragraph{Future Work}

Several directions merit further investigation. Extending the pattern library to include additional laundering typologies would broaden Tide's applicability. Disaggregating model performance by pattern type would clarify which architectures are best suited for detecting specific laundering behaviours.

The integration of Temporal Graph Networks (TGNs), which explicitly model time-evolving graph structure, represents a promising avenue given Tide's emphasis on temporal dynamics. Additionally, incorporating adversarial pattern generation, where laundering schemes evolve in response to detection models, could produce more challenging benchmarks that better simulate the adaptive nature of financial crime.

Future work could also explore cross-institutional scenarios with multiple financial networks, enabling research on information-sharing protocols and federated detection approaches. Finally, establishing a community-driven pattern repository, where researchers contribute new typologies with formal structural and temporal specifications, would extend the generator's utility beyond the patterns we have implemented.

\bibliographystyle{ACM-Reference-Format}
\bibliography{references}

\section*{Appendix}

\subsection{Entity Attribute Specifications}
\label{app:entity-attributes}

This section provides detailed attribute definitions for each entity type in the Tide graph model. Tables~\ref{tab:individual_node_attributes}--\ref{tab:fi_node_attributes} specify the type-specific attributes that complement the common attributes defined in Section~4.2.

\begin{table}[h!]
\centering
\begin{tabularx}{\textwidth}{l X l}
\toprule
\textbf{Attribute} & \textbf{Description} & \textbf{Type} \\
\midrule
\textbf{Name} & An individual's full name. & String \\
\textbf{Age group} & The individual's age group. & Categorical \\
\textbf{Occupation} & The individual's occupation. & Categorical \\
\textbf{Gender} & The individual's gender. & Categorical \\
\bottomrule
\end{tabularx}
\caption{Specific attributes for nodes with the \textit{individual} entity type.}
\label{tab:individual_node_attributes}
\end{table}

\begin{table}[h!]
\centering
\begin{tabularx}{\textwidth}{l X l}
\toprule
\textbf{Attribute} & \textbf{Description} & \textbf{Type} \\
\midrule
\textbf{business\_category} & The primary sector or type of business activity. & Categorical \\
\textbf{incorporation\_year} & The year the business was legally incorporated. & Integer \\
\textbf{number\_of\_employees} & Approximate number of people employed by the business. & Integer \\
\textbf{is\_high\_risk\_category} & Flag for businesses in a high-risk category. & Boolean \\
\bottomrule
\end{tabularx}
\caption{Specific attributes for nodes with the \textit{business} entity type.}
\label{tab:business_node_attributes}
\end{table}

\begin{table}[h!]
\centering
\begin{tabularx}{\textwidth}{l X l}
\toprule
\textbf{Attribute} & \textbf{Description} & \textbf{Type} \\
\midrule
\textbf{account\_category} & Type of account (e.g., Current, Savings, Cash). & Categorical \\
\textbf{currency} & The currency used by the account. & String \\
\textbf{owner\_id} & ID of the Individual or Business that owns this account. & String \\
\textbf{institution\_id} & ID of the Financial Institution that hosts this account. & String \\
\bottomrule
\end{tabularx}
\caption{Specific attributes for nodes with the \textit{account} entity type.}
\label{tab:account_node_attributes}
\end{table}

\begin{table}[h!]
\centering
\begin{tabularx}{\textwidth}{l X l}
\toprule
\textbf{Attribute} & \textbf{Description} & \textbf{Type} \\
\midrule
\textbf{institution\_name} & Name of the financial institution. & String \\
\textbf{country} & Country of legal registration or regulation. & String \\
\bottomrule
\end{tabularx}
\caption{Specific attributes for nodes with the \textit{financial institution} entity type.}
\label{tab:fi_node_attributes}
\end{table}

\subsection{Risk Scoring and Pattern Parameters}
\label{app:risk-patterns}

This section documents the default parameters used for entity risk scoring and pattern generation. Table~\ref{tab:risk-weights} presents the risk factor weights used in the entity risk scoring framework described in Section~4.2.1. Tables~\ref{tab:pattern_overview} and~\ref{tab:background_overview} provide an overview of the implemented laundering and background activity patterns, respectively.

\begin{table}[t]
  \caption{Default risk factor weights used in entity risk scoring.}
  \label{tab:risk-weights}
  \centering
  \begin{tabular}{llr}
    \toprule
    \textbf{Entity Type} & \textbf{Risk Factor} & \textbf{Weight ($w_i$)} \\
    \midrule
    \multirow{3}{*}{Individual}
      & Base risk                      & 0.05 \\
      & High-risk age group            & 0.15 \\
      & High-risk occupation           & 0.12 \\
    \midrule
    \multirow{4}{*}{Business}
      & Base risk                      & 0.10 \\
      & Cash-intensive business category & 0.25 \\
      & Very small company ($\leq$5 employees) & 0.10 \\
    \midrule
    Both & High-risk jurisdiction         & 0.20 \\
    \bottomrule
  \end{tabular}
\end{table}

\begin{table}[t]
\centering
\caption{Overview of implemented laundering patterns.}
\label{tab:pattern_overview}
\begin{tabular}{@{}lllc@{}}
\toprule
\textbf{Pattern} & \textbf{Topology} & \textbf{Timing} & \textbf{Additional layering} \\
\midrule
Overseas Transfers & 1 → n overseas (n=2–5) & Periodic (7/14/30d) & \checkmark \\
Rapid Movement & n → 1 → cash-out (n=2–7) & Burst (1–24h delay) & \checkmark \\
Front Business & Cash → business → overseas & Paired (0.5–6h) & \checkmark \\
U-Turn & Source → chain → return & Sequential (1–5d/hop) & Inherent \\
Synchronised & n → 1 recipient (n=3–8) & Coordinated (2h window) & — \\
\bottomrule
\end{tabular}
\end{table}

\begin{table}[t]
\centering
\caption{Overview of background activity patterns. Baseline patterns model routine financial behaviour; counter-leakage patterns introduce legitimate activity that shares features with fraud patterns.}
\label{tab:background_overview}
\begin{tabular}{@{}llll@{}}
\toprule
\textbf{Pattern} & \textbf{Group} & \textbf{Fraud feature countered} \\
\midrule
Random everyday payments     & Baseline & --- \\
High-value transactions      & Baseline & --- \\
Salary payments              & Baseline & --- \\
Fraudster background         & Baseline & Behavioural anomaly \\
\midrule
Legitimate bursts            & Counter-leakage & Temporal clustering \\
Legitimate periodic          & Counter-leakage & Periodic timing \\
Legitimate chains            & Counter-leakage & Multi-hop topology \\
Legitimate rapid flow        & Counter-leakage & Inflow--outflow sequences \\
Legitimate cash operations   & Counter-leakage & Cash transaction types \\
Legitimate structuring       & Counter-leakage & Near-threshold amounts \\
High-risk entity activity    & Counter-leakage & Risk-score correlation \\
\bottomrule
\end{tabular}
\end{table}

\subsection{Feature Engineering and Export}
\label{app:features}

This section details the features included in exported datasets and the preprocessing configurations used for baseline models. Table~\ref{tab:exported_features} lists all node and edge features available in the exported dataset, with categorical features one-hot encoded. Internal generation attributes (e.g., \texttt{risk\_score}, \texttt{occupation}, \texttt{is\_high\_risk\_category}) are excluded to prevent data leakage. Table~\ref{tab:gfp-config} presents the Graph Feature Preprocessor (GFP) configuration adopted from Altman et al.~\cite{altman2023realistic}.

\begin{table}[t]
\centering
\caption{Node and edge features included in the exported dataset. Categorical features are one-hot encoded. Internal generation attributes (\texttt{risk\_score}, \texttt{occupation}, \texttt{is\_high\_risk\_category}, etc.)\ are excluded.}
\label{tab:exported_features}
\begin{subtable}[t]{0.48\textwidth}
\centering
\caption{Node features.}
\label{tab:node_features}
\begin{tabular}{@{}lll@{}}
\toprule
\textbf{Feature} & \textbf{Type} & \textbf{Encoding} \\
\midrule
\texttt{node\_type}          & Cat. & One-hot (4) \\
\texttt{country\_code}       & Cat. & One-hot (top 30) \\
\texttt{account\_category}   & Cat. & One-hot (5) \\
\texttt{gender}              & Cat. & One-hot (2) \\
\texttt{age\_group}          & Cat. & One-hot (5) \\
\texttt{business\_category}  & Cat. & One-hot (top 20) \\
\texttt{incorporation\_year} & Num. & Raw \\
\texttt{creation\_year}      & Num. & Raw \\
\bottomrule
\end{tabular}
\end{subtable}
\hfill
\begin{subtable}[t]{0.48\textwidth}
\centering
\caption{Edge features (transaction edges).}
\label{tab:edge_features}
\begin{tabular}{@{}lll@{}}
\toprule
\textbf{Feature} & \textbf{Type} & \textbf{Encoding} \\
\midrule
\texttt{amount}              & Num. & Raw \\
\texttt{timestamp}           & Num. & Unix seconds \\
\texttt{time\_since\_prev}   & Num. & Seconds \\
\texttt{transaction\_type}   & Cat. & One-hot (5) \\
\texttt{currency}            & Cat. & One-hot \\
\bottomrule
\end{tabular}
\end{subtable}
\end{table}

\begin{table}[h]
  \caption{Graph Feature Preprocessor (GFP) configuration, adopted from Altman et al.~\cite{altman2023realistic}.}
  \label{tab:gfp-config}
  \centering
  \begin{tabular}{ll}
    \toprule
    Parameter & Value \\
    \midrule
    Batch size & 128 \\
    Patterns & Scatter-gather, temporal cycles, \\
             & simple cycles, vertex statistics \\
    Scatter-gather window & 6 hours \\
    Cycle / vertex stats window & 1 day \\
    Max simple cycle length & 10 \\
    Vertex statistics fields & Amount, Timestamp \\
    \bottomrule
  \end{tabular}
\end{table}

\subsection{Calibration Data Sources}
\label{app:calibration}

Transaction amount and frequency distributions in Tide were calibrated using publicly available reference data. Table~\ref{tab:frps-raw} presents the noncash payment volumes from the 2022 Federal Reserve Payments Study~\cite{frps2022}, which informed our transaction type frequency distributions as described in Section~4.5.2.

\begin{table}[htbp!]
\centering\small
\caption{FRPS 2021 noncash payment volumes by category.}
\label{tab:frps-raw}
\begin{tabular}{lrr}
\toprule
FRPS Category & Volume (B) & Share \\
\midrule
Non-prepaid debit cards & 87.8 & 42.2\% \\
Credit cards            & 51.1 & 24.6\% \\
ACH debit transfers     & 20.3 & 9.8\%  \\
Prepaid debit cards     & 18.1 & 8.7\%  \\
ACH credit transfers    & 15.9 & 7.6\%  \\
Checks                  & 11.2 & 5.4\%  \\
ATM withdrawals         & 3.7  & 1.8\%  \\
\midrule
Total                   & 208.1 & 100\% \\
\bottomrule
\end{tabular}
\end{table}

\subsection{Extended Experimental Results}
\label{app:results}

This section provides the full performance metrics that complement the summary results presented in Section~5.2.3. Table~\ref{tab:model_comparison} reports detailed results for the primary experimental condition (low illicit ratio, 0.10\% class imbalance). All metrics are computed at Youden's optimal threshold, with results reported as mean $\pm$ standard deviation over 5 runs.

\begin{table}[t]
\centering
\caption{Hyperparameter configuration for GBT models.}
\label{tab:gbt_hyperparameters}
\begin{tabular}{@{}lcc@{}}
\toprule
\textbf{Parameter} & \textbf{LightGBM} & \textbf{XGBoost} \\
\midrule
\texttt{n\_estimators}     & 1000      & 1000      \\
\texttt{max\_depth}        & 6         & 6         \\
\texttt{num\_leaves}       & 64        & --        \\
\texttt{learning\_rate}    & 0.01      & 0.01      \\
\texttt{reg\_lambda} (L2)  & 5.0       & 5.0       \\
\texttt{reg\_alpha} (L1)   & 1.5       & 0.5       \\
\texttt{subsample}         & 0.8       & 0.8       \\
\texttt{colsample\_bytree} & 0.8       & 0.8       \\
\texttt{min\_child\_samples/weight} & 50 & 5       \\
class weighting\textsuperscript{$\dagger$} & balanced & 200 \\
\texttt{gamma}             & --        & 0.2       \\
\bottomrule
\end{tabular}

\vspace{2mm}
\parbox{\linewidth}{\footnotesize $\dagger$ LightGBM: \texttt{class\_weight='balanced'} (auto-computed); XGBoost: \texttt{scale\_pos\_weight=200}.}
\end{table}

\begin{table}[t]
\centering
\caption{Hyperparameter configuration for GNN models.}
\label{tab:gnn_hyperparameters}
\begin{tabular}{@{}lccc@{}}
\toprule
\textbf{Parameter} & \textbf{GIN} & \textbf{GIN+EU} & \textbf{PNA} \\
\midrule
hidden dimension      & 64   & 64   & 64   \\
number of layers      & 4    & 4    & 2    \\
dropout               & 0.2  & 0.2  & 0.1  \\
final dropout         & 0.5  & 0.5  & 0.5  \\
learning rate         & \multicolumn{3}{c}{0.001} \\
weight decay          & \multicolumn{3}{c}{0.0005} \\
minority class weight & \multicolumn{3}{c}{50} \\
\bottomrule
\end{tabular}
\end{table}

\begin{table}[htbp!]
\centering
\caption{Comparison of model performance on fraud detection. Homophily index: 0.0023. Class imbalance of 0.10\%. Lift = PR-AUC / fraud\_rate (times better than random). All metrics (F1, Precision, Recall) computed at \textbf{Youden's optimal threshold}. Results reported as mean $\pm$ std over 5 runs.}
\label{tab:results_li}
\renewcommand{\arraystretch}{1.2}
\setlength{\tabcolsep}{5pt}
\begin{tabular}{lcccccc|c}
\toprule
\textbf{Model} & \textbf{F1} & \textbf{Precision} & \textbf{Recall} & \textbf{PR-AUC} & \textbf{Lift} & \textbf{Youden's J} & \textcolor{gray}{\textbf{ROC-AUC}} \\
\midrule
GIN & $5.95 \pm 1.27$ & $3.07 \pm 0.67$ & $94.33 \pm 2.62$ & $69.86 \pm 6.55$ & $670.63 \pm 62.91$ & $90.66 \pm 3.38$ & \textcolor{gray}{$97.64 \pm 1.05$} \\
\rowcolor{gray!15}
GIN+EU & $4.45 \pm 3.09$ & $2.31 \pm 1.66$ & $87.55 \pm 7.82$ & $39.43 \pm 20.99$ & $378.46 \pm 201.54$ & $80.91 \pm 10.56$ & \textcolor{gray}{$94.66 \pm 3.99$} \\
PNA & \textbf{$69.90 \pm 0.00$} & \textbf{$57.86 \pm 0.00$} & $88.26 \pm 0.00$ & $69.20 \pm 0.00$ & $664.30 \pm 0.00$ & $88.19 \pm 0.00$ & \textcolor{gray}{$98.44 \pm 0.00$} \\
LightGBM+GFP & $12.13 \pm 7.18$ & $6.64 \pm 4.20$ & $92.81 \pm 8.64$ & $21.45 \pm 13.23$ & $205.95 \pm 127.04$ & $90.12 \pm 10.92$ & \textcolor{gray}{$96.60 \pm 5.38$} \\
XGBoost+GFP & $10.19 \pm 2.04$ & $5.38 \pm 1.15$ & \textbf{$99.31 \pm 0.40$} & $35.45 \pm 2.59$ & $340.28 \pm 24.88$ & \textbf{$97.17 \pm 0.20$} & \textcolor{gray}{$99.77 \pm 0.02$} \\
LightGBM & $8.01 \pm 2.21$ & $4.20 \pm 1.20$ & $95.73 \pm 3.50$ & \textbf{$78.05 \pm 3.61$} & \textbf{$749.19 \pm 34.64$} & $92.80 \pm 3.16$ & \textcolor{gray}{$98.80 \pm 1.75$} \\
XGBoost & $4.22 \pm 0.38$ & $2.16 \pm 0.20$ & $97.04 \pm 0.56$ & $72.01 \pm 0.50$ & $691.22 \pm 4.82$ & $91.81 \pm 0.22$ & \textcolor{gray}{$99.42 \pm 0.05$} \\
\bottomrule
\end{tabular}
\end{table}

\begin{table}[htbp!]
\centering
\caption{Comparison of model performance on fraud detection. Homophily index:  0.0035. Class imbalance of 0,19\%. Lift = PR-AUC / fraud\_rate (times better than random). All metrics (F1, Precision, Recall) computed at \textbf{Youden's optimal threshold}. Results reported as mean $\pm$ std over 5 runs.}
\label{tab:results_hi}
\renewcommand{\arraystretch}{1.2}
\setlength{\tabcolsep}{5pt}
\begin{tabular}{lcccccc|c}
\toprule
\textbf{Model} & \textbf{F1} & \textbf{Precision} & \textbf{Recall} & \textbf{PR-AUC} & \textbf{Lift} & \textbf{Youden's J} & \textcolor{gray}{\textbf{ROC-AUC}} \\
\midrule
GIN & $9.83 \pm 2.35$ & $5.20 \pm 1.29$ & $91.65 \pm 5.28$ & $71.14 \pm 8.70$ & $378.40 \pm 46.29$ & $87.98 \pm 6.14$ & \textcolor{gray}{$96.71 \pm 1.90$} \\
\rowcolor{gray!15}
GIN+EU & $4.88 \pm 2.05$ & $2.52 \pm 1.09$ & $86.59 \pm 5.20$ & $48.42 \pm 17.05$ & $257.56 \pm 90.71$ & $78.49 \pm 7.27$ & \textcolor{gray}{$93.61 \pm 2.85$} \\
PNA & \textbf{$31.67 \pm 13.94$} & \textbf{$20.20 \pm 10.83$} & $91.42 \pm 1.34$ & $75.63 \pm 5.85$ & $402.30 \pm 31.14$ & $90.31 \pm 0.57$ & \textcolor{gray}{$98.61 \pm 0.22$} \\
LightGBM+GFP & $20.98 \pm 10.06$ & $12.13 \pm 6.46$ & $97.35 \pm 1.84$ & $64.40 \pm 24.36$ & $342.56 \pm 129.57$ & $95.28 \pm 2.21$ & \textcolor{gray}{$99.11 \pm 0.94$} \\
XGBoost+GFP & $12.46 \pm 0.65$ & $6.65 \pm 0.37$ & \textbf{$97.90 \pm 0.28$} & $32.20 \pm 1.68$ & $171.26 \pm 8.93$ & $95.06 \pm 0.12$ & \textcolor{gray}{$99.53 \pm 0.03$} \\
LightGBM & $28.59 \pm 8.68$ & $17.04 \pm 5.72$ & $96.69 \pm 2.31$ & $70.47 \pm 27.91$ & $374.85 \pm 148.46$ & $95.51 \pm 2.97$ & \textcolor{gray}{$99.45 \pm 0.68$} \\
XGBoost & $21.82 \pm 1.78$ & $12.30 \pm 1.13$ & $97.48 \pm 0.17$ & \textbf{$85.12 \pm 0.52$} & \textbf{$452.78 \pm 2.76$} & \textbf{$96.03 \pm 0.04$} & \textcolor{gray}{$99.80 \pm 0.01$} \\
\bottomrule
\end{tabular}
\end{table}

Table~\ref{tab:model_comparison} compares the performance of PNA and XGBoost models.

\begin{table}[t]
\centering
\caption{Performance comparison of PNA and XGBoost models on the classification task on the \textbf{HI} dataset. Precision-Recall curves show Average Precision (AP) and F1 scores at optimal thresholds.}
\label{tab:model_comparison}
\includegraphics[width=0.9\columnwidth]{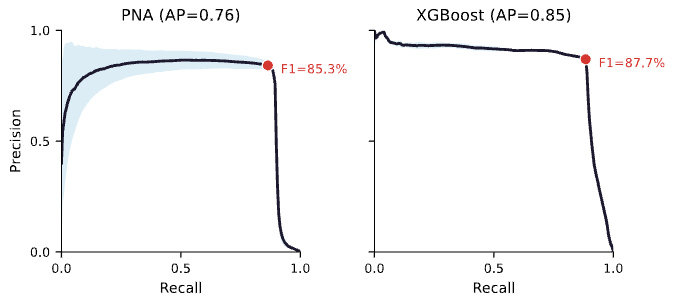}
\end{table}

\subsection{Scalability Analysis}
\label{app:scalability}

To evaluate the scalability of Tide's graph generation, we conducted experiments across six different graph sizes, ranging from 1,000 to 12,000 individuals. The results demonstrate good scalability, with all configurations successfully generating the specified patterns. However, an analysis of memory usage reveals that while the system scales efficiently from a complexity standpoint, absolute memory consumption becomes a practical bottleneck on resource-constrained systems, which impacts the overall generation time.

Table~\ref{tab:h2_results_aggregated_compact} presents the performance metrics for these experiments, including the peak memory usage for each configuration.

\begin{table}[htbp]
\centering
\small
\caption{Aggregated generation performance across different graph scales on an M2 MacBook Air with 8\,GB of RAM. Results are presented as mean $\pm$ standard deviation over 3 runs.}
\label{tab:h2_results_aggregated_compact}
\begin{tabular}{l r r r r}
\toprule
\textbf{Configuration} & \multicolumn{1}{c}{\textbf{Transactions}} & \multicolumn{1}{c}{\textbf{Total ($N$)}} & \multicolumn{1}{c}{\textbf{Wall-time}} & \multicolumn{1}{c}{\textbf{Peak Memory}} \\
& \multicolumn{1}{c}{\textbf{[millions]}} & \multicolumn{1}{c}{\textbf{[millions]}} & \multicolumn{1}{c}{\textbf{[s]}} & \multicolumn{1}{c}{\textbf{[MB]}} \\
\midrule
Test-1k  & $0.372 \pm 0.004$  & $0.377 \pm 0.004$  & $2.3 \pm 0.3$   & $490 \pm 30$    \\
Test-2k       & $0.948 \pm 0.005$  & $0.957 \pm 0.005$  & $5.2 \pm 0.3$   & $1090 \pm 30$   \\
Test-3.5k & $1.99 \pm 0.01$    & $2.00 \pm 0.01$    & $11.0 \pm 0.9$  & $2200 \pm 50$   \\
Test-5k       & $3.28 \pm 0.03$    & $3.30 \pm 0.03$    & $18 \pm 1$      & $2700 \pm 200$  \\
Test-8k & $5.99 \pm 0.02$    & $6.03 \pm 0.02$    & $43 \pm 3$      & $3600 \pm 40$   \\
Test-12k        & $10.17 \pm 0.04$   & $10.23 \pm 0.04$   & $120 \pm 10$    & $3700 \pm 100$  \\
\bottomrule
\end{tabular}
\end{table}

\begin{figure}[htbp]
    \centering
    \begin{subfigure}[t]{0.48\linewidth}
        \centering
        \includegraphics[width=\linewidth]{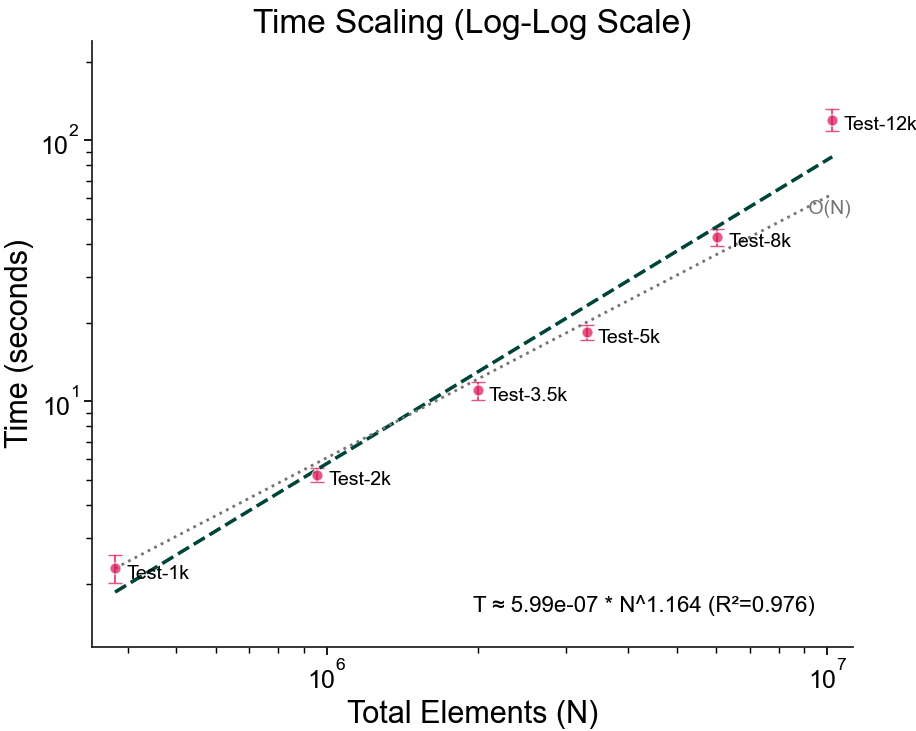}
        \caption{Generation time vs total graph elements ($N$).}
        \label{fig:time_complexity}
    \end{subfigure}
    \hfill
    \begin{subfigure}[t]{0.48\linewidth}
        \centering
        \includegraphics[width=\linewidth]{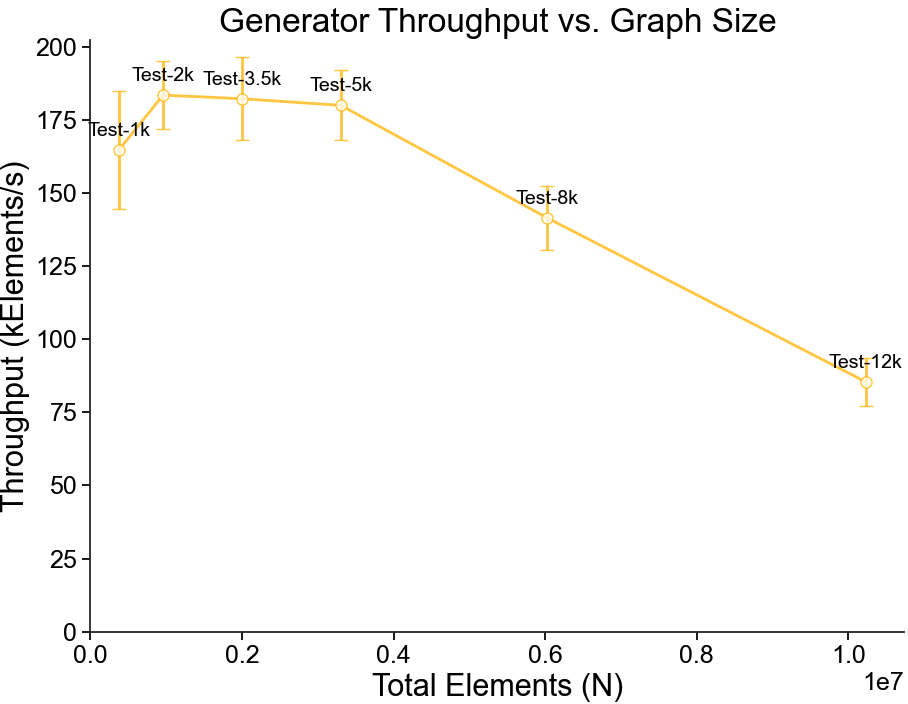}
        \caption{Throughput in kElements/s vs total graph elements.}
        \label{fig:throughput}
    \end{subfigure}
    
    \vspace{0.5cm}
    
    \begin{subfigure}[t]{0.48\linewidth}
        \centering
        \includegraphics[width=\linewidth]{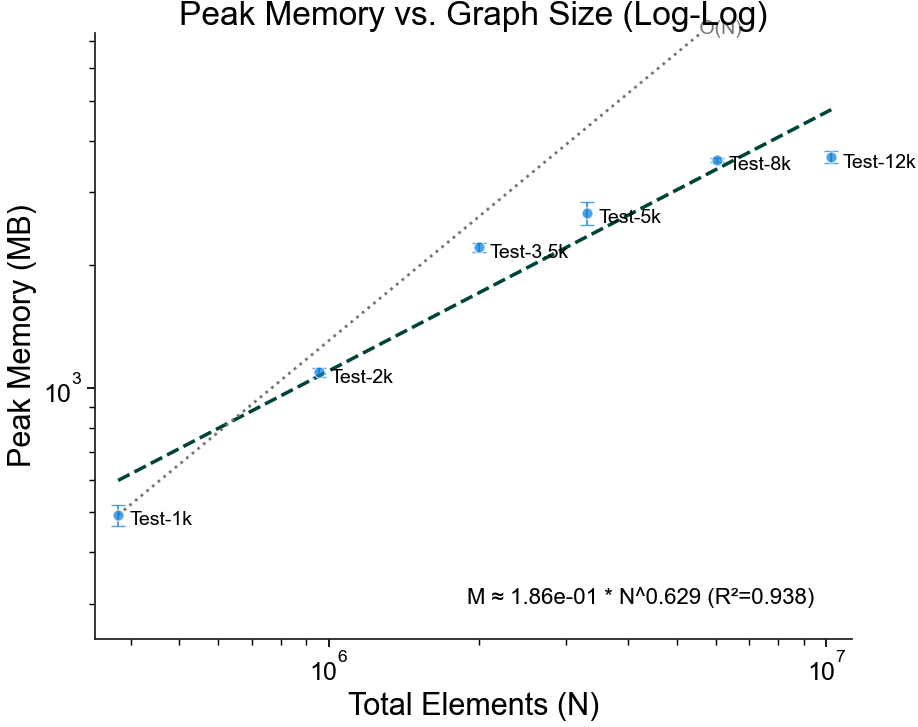}
        \caption{Peak memory usage vs total graph elements ($N$).}
        \label{fig:memory_scaling}
    \end{subfigure}
    \hfill
    \begin{subfigure}[t]{0.48\linewidth}
        \centering
        \includegraphics[width=\linewidth]{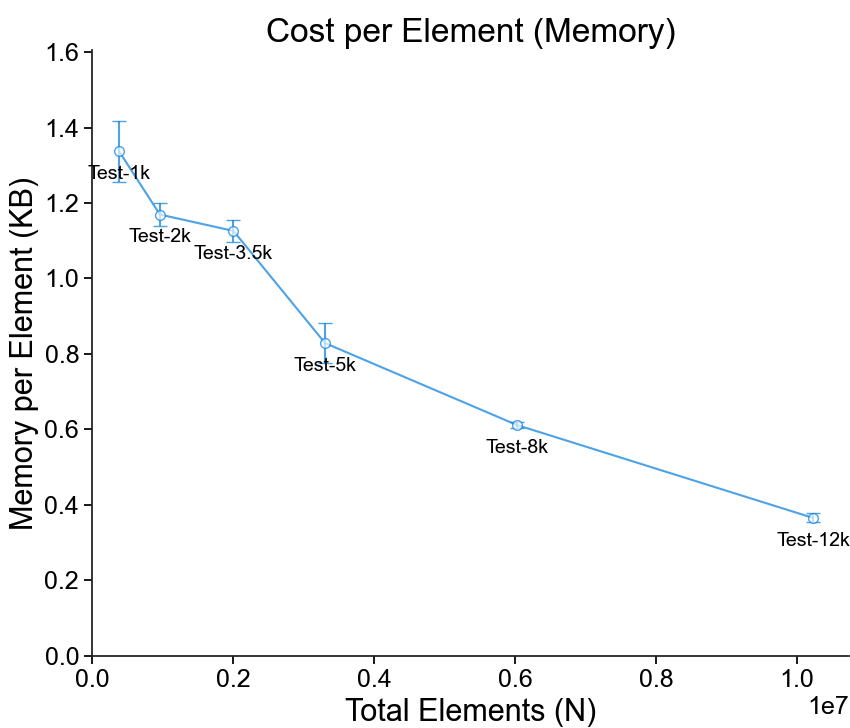}
        \caption{Additional memory per graph element.}
        \label{fig:memory_per_element}
    \end{subfigure}
    
    \caption{Performance analysis of the graph generator. (a) Log-log plot showing generation time versus total graph elements with near-linear scaling (exponent $\alpha = 1.164$, $R^2 = 0.976$). (b) Average throughput in kElements/s. (c) Peak memory usage scaling with sub-linear exponent of 0.627. (d) Memory efficiency showing additional memory required per element. Gray dotted lines in (a) and (c) represent $\mathcal{O}(N)$ for reference.}
    \label{fig:performance_analysis}
\end{figure}

Figure~\ref{fig:performance_analysis} illustrates the performance characteristics of the generator across the different scales.

As shown in Table~\ref{tab:h2_metrics}, all metrics for the generation time met their predefined success thresholds. We calculated a scaling exponent of $\alpha=1.164$, which indicates a near-linear time complexity of $O(N^{1.164})$ with graph size $N$. This value is well within our success criterion of $\alpha \leq 1.5$. The coefficient of determination $R^2 = 0.976$ shows that the system's performance remains predictable across different scales. This strong, near-linear relationship between graph size and generation time is also supported by a Pearson correlation of $r=0.970$ and a statistically significant $p$-value of 0.00137.

\begin{table}[htbp]
\centering
\small
\caption{Statistical metrics from scalability analysis.}
\label{tab:h2_metrics}
\begin{tabular}{l r r}
\toprule
\textbf{Metric}                            & \textbf{Value} & \textbf{Target} \\
\midrule
Pearson correlation ($r$)                  & 0.970          & $> 0.9$                    \\
$p$-value                                  & 0.00137        & $< 0.05$                   \\
Scaling exponent ($\alpha$)                & 1.164          & $\alpha \leq 1.5$          \\
Coefficient of determination ($R^2$)       & 0.976          & $> 0.9$                    \\
\bottomrule
\end{tabular}
\end{table}

The analysis of peak memory usage reveals a scaling behaviour of $O(N^{0.627})$. This indicates that the memory required per element decreases as the graph grows. Figure~\ref{fig:memory_per_element} shows the additional memory required per graph element, confirming this trend.

The generator's throughput, shown in Figure~\ref{fig:throughput}, further illustrates the practical performance characteristics. Throughput peaks for moderately-sized graphs at around 185 kElements/s before declining as the graph size grows. This trend indicates that beyond a certain point, the overhead associated with memory pressure begins to outpace the raw processing capability, which reinforces the conclusion that memory access, rather than computation, becomes the primary bottleneck for larger-scale generation on resource-constrained systems.

The experiments were run on a machine with only 8\,GB of RAM, where the operating system itself consumes around 5\,GB. This leaves limited memory for the generation process. As the graph size increases, the system experiences increased pressure on memory resources, likely leading to swapping and other overhead that inflates the wall-clock time.

Despite the sub-linear memory scaling, the $O(N^{1.164})$ scaling in generation time is still affected by a memory bottleneck in practice. As seen in Table~\ref{tab:h2_results_aggregated_compact}, the absolute memory required for generation is significant, reaching approximately 3.7\,GB for a graph with over 10 million transactions.

\end{document}